\PassOptionsToPackage{table, x11names}{xcolor}

\documentclass{article} %

\usepackage{colm2024_conference}

\usepackage{graphicx}
\usepackage{subcaption}
\usepackage{microtype}
\usepackage{hyperref}
\usepackage{url}
\usepackage{booktabs}
\usepackage{cellspace}
\usepackage{multirow}
\usepackage{tcolorbox}
\usepackage{enumitem}
\usepackage{cleveref}
\usepackage{wrapfig}
\usepackage{xcolor}
\usepackage{tablefootnote}

\definecolor{darkblue}{rgb}{0, 0, 0.5}
\hypersetup{colorlinks=true, citecolor=darkblue, linkcolor=darkblue, urlcolor=darkblue}

\makeatletter
\AtBeginDocument{%
\renewcommand{\sectionautorefname}{\S\@gobble}

\renewcommand{\subsectionautorefname}{\S\@gobble} 
\renewcommand{\subsubsectionautorefname}{\S\@gobble} 
}
\makeatother

\colmfinaltrue

\title{How to Train Long-Context Language Models (Effectively)}

\author{Tianyu Gao\thanks{Equal contribution and corresponding authors.} $\;$ Alexander Wettig$^*$ $\;$ Howard Yen $\;$ Danqi Chen \\
   Princeton Language and Intelligence, Princeton University\\
   \texttt{\{tianyug,awettig,hyen,danqic\}@cs.princeton.edu} \\
}

\begin{document}

\providecommand{\todo}[1]{{\protect\color{cyan}{[TODO: #1]}}}
\providecommand{\danqi}[1]{{\protect\color{orange}{[Danqi: #1]}}}
\providecommand{\tianyu}[1]{{\protect\color{blue}{[Tianyu: #1]}}}
\providecommand{\howard}[1]{{\protect\color{purple}{[Howard: #1]}}}
\providecommand{\alex}[1]{{\protect\color{green!60!black}{[Alex: #1]}}}

\providecommand{\revision}[1]{{#1}}
\newcommand{\alexedit}[1]{{\protect{#1}}}
\newcommand{\tgedit}[1]{{\protect{#1}}}

\providecommand{\todo}[1]{{\protect\color{red}{}}}
\providecommand{\danqi}[1]{{\protect\color{orange}{}}}
\providecommand{\tianyu}[1]{{\protect\color{blue}{}}}
\providecommand{\howard}[1]{{\protect\color{purple}{}}}

\newcommand{\arrawedit}[1]{{\protect{#1}}}
\newcommand{\arrtgedit}[1]{{\protect{#1}}}

\newcommand{\cmark}{\ding{51}}
\newcommand{\xmark}{\ding{55}}

\newcommand\ti[1]{\textit{#1}}
\newcommand\ts[1]{\textsc{#1}}
\newcommand\tf[1]{\textbf{#1}}
\newcommand\ttt[1]{\texttt{#1}}
\newcommand\mf[1]{\mathbf{#1}}
\newcommand\tmp[1]{\color{gray}{#1}}
\newcommand\warn[1]{\textbf{\color{red}{#1}}}
\newcommand\mr[1]{\mathrm{#1}}
\newcommand\mc[1]{\mathcal{#1}}

\newcommand\myeq{\stackrel{\mathclap{\normalfont\mbox{i.i.d.}}}{~}}
\providecommand{\todon}{
    {\protect\color{red}{00.00}}
}

\newcommand{\identical}{identical}
\newcommand{\Identical}{Identical}

\newcommand{\la}{$_\texttt{large}$}
\newcommand{\ba}{$_\texttt{base}$}

\renewcommand{\paragraph}[1]{\vspace{0.2cm}\noindent\textbf{#1}}
\newcommand{\tpf}[1]{\noindent\textbf{#1}}
\newcommand{\tableindent}{~~}

\newcommand{\vani}{{\sc{Vanilla}}}
\newcommand{\replug}{{\sc{RePlug}}}

\newcommand{\prevdoc}{{\sc{PrevDoc}}}
\newcommand{\retdoc}{RetDoc}

\newcommand{\ccc}{\textasciicircum}

\newcommand{\tableskip}{\noalign{\vskip 2pt}}
\newcommand{\headercolor}{\rowcolor{gray!15}}
\newcommand{\headercolorlong}{\rowcolor{gray!17}}

\newcommand{\zs}{Zero{\sc{Scrolls}}}

\newcommand{\citedb}[1]{{\color{darkblue}{#1}}}

\newcommand{\rpfilter}{RP$_\text{train-filter}$}
\newcommand{\rpcat}{RP$_\text{train-cat}$}

\newcommand{\menc}{\mathcal{M}_\textrm{enc}}
\newcommand{\mdec}{\mathcal{M}_\textrm{dec}}
\newcommand{\concat}{\textsc{concat}}

\newcommand{\denc}{d_\textrm{enc}}
\newcommand{\ddec}{d_\textrm{dec}}

\newcommand{\jsonkv}{Recall}
\newcommand{\rag}{RAG}
\newcommand{\rerank}{Re-rank}
\newcommand{\icl}{ICL}
\newcommand{\qa}{QA}
\newcommand{\summ}{Summ.}
\newcommand{\summfull}{Summarization}
\newcommand{\avg}{Avg.}
\newcommand{\shortmix}{ShortMix}
\newcommand{\ours}{ProLong}
\newcommand{\inst}{supervised fine-tuning} %
\newcommand{\Inst}{Supervised fine-tuning} %
\newcommand{\INST}{Supervised Fine-Tuning} %
\newcommand{\sft}{SFT}
\newcommand{\helmet}{HELMET}

\newcommand{\llama}{Llama-3-8B}
\newcommand{\llamabase}{Llama-3-8B-Base}
\newcommand{\llamainst}{Llama-3-8B-Instruct}

\newcommand{\nocha}{NoCha}
\newcommand{\ctraining}{\tgedit{continued} training}

\maketitle

\begin{abstract}

We study \ctraining{} and \inst{} (\sft{}) of a language model (LM) to make effective use of long-context information.
We first establish a reliable evaluation protocol to guide model development---instead of perplexity or simple needle-in-a-haystack (NIAH) tests,
we use \alexedit{a broad set of} long-context \arrawedit{downstream} \alexedit{tasks}, and we evaluate \alexedit{models} after \sft{}
as  this better reveals long-context abilities.
Supported by our robust evaluations, we run thorough experiments to decide the data mix \alexedit{for continued pre-training}, the instruction tuning dataset, and many other design choices \arrawedit{such as position extrapolation}.
 We find that 
(1) code repositories and books are excellent sources of long data, but it is crucial to combine them with high-quality short\arrawedit{-context} data; 
(2) \alexedit{training with a sequence length beyond the evaluation length boosts long-context performance};
(3) for \sft{}, using only short instruction datasets yields strong performance  on long-context tasks.
Our final model, \textbf{\ours{}-8B}, \alexedit{which is initialized from Llama-3 and trained on 40B tokens,}
demonstrates state-of-the-art long-context performance among similarly sized models
at a length of 128K. %
\ours{} outperforms Llama-3.1-8B-Instruct on the majority of long-context tasks despite \arrawedit{using} only 5\% as many tokens during long-context training.
Additionally, \ours{} can effectively process up to 512K tokens, one of the longest context windows of publicly available LMs.\footnote{Our code, data, and models are available at  \url{https://github.com/princeton-nlp/ProLong}.}

\end{abstract}

\begin{figure}[h!]
    \centering
\begin{tcolorbox}[title={\bf Takeaways for \tgedit{continued} training of long-context models}, colback={orange!05!white},colframe={orange!30!white},coltitle=black]

    \begin{itemize}[left=0pt]

        \item \tf{Evaluation} (\autoref{sec:eval}):  
        We \alexedit{target a range of long-context downstream tasks} instead of perplexity or needle-in-a-haystack, while checking if the short-context performance is preserved. 
        We evaluate models after \sft{}, which produces a clearer  signal on long-context tasks.%

        \item \tf{Data engineering} (\autoref{sec:data}): 
        We conduct a series of ablations at a 5B-token scale. 
        We find that 
       using  code repositories and long books as long-context data and 
       mixing them with high-quality short-context data is crucial  for both long-context performance and retaining the short-context capabilities of the pre-trained model.  %

        \item \tf{Scaling the data and the length} (\autoref{sec:howlong}):
        We scale up the training to 20B tokens at a 64K training length and 20B tokens at a 512K training length.
        Surprisingly, training on  contexts longer than the evaluation length yields additional benefits. %
        
       \item \tf{\Inst{}} (\autoref{sec:sft}): 
       We find that \sft{} with standard, short-context instruction datasets is  sufficient for achieving good performance. 
       Contrary to previous study, 
       long synthetic instruction data does not boost the result in our setting.
       \item \tf{\ours{} models} (\autoref{sec:recipe}): We present our final recipe and evaluation results here. \alexedit{All our code, data, and models are made publically available.}
    \end{itemize}
\end{tcolorbox}
\label{fig:takeaways}
\vspace{-10pt}
\end{figure}

\section{Introduction}

The ability of language models (LMs) to process extremely long inputs (for example, 128K tokens) has enabled new applications, such as book summarization or learning new tasks on the fly from many examples. %
However, adapting
LMs to process long contexts is challenging from an infrastructure and data perspective, and many design decisions are not well understood by open-source practitioners.

While many works have focused on extending the context length of pre-trained LMs with minimal training \citep[\arrawedit{\textit{inter alia}}]{chen2023extending,peng2024yarn},
the above methods \arrawedit{fail to solve even} the simple needle-in-a-haystack \citep[NIAH;][]{gkamradt_llmtest_needleinahaystack_2024} task
and  it is necessary to \arrawedit{continue LLM training on billions of tokens of long documents to learn this task robustly}. %
Frontier open-source models, such as Llama-3.1~\citep{dubey2024llama} and Jamba~\citep{team2024jamba15}, also employ a long-context \ctraining{} stage, followed by \inst{} (\sft{}) on instruction data. 
We adopt the same setting and study \ctraining{} and \sft{} of a  pre-trained LM for effective long-context use.

We first establish a reliable evaluation protocol to provide a meaningful signal for model development. 
Most existing works rely on either perplexity or NIAH  for ablating training recipes.
We demonstrate that neither is robust for guiding the development 
and opt for a broad range of downstream applications, such as retrieval-augmented generation (RAG), long-document summarization, and many-shot in-context learning (ICL). 
Importantly, we also \alexedit{conduct our evaluations after performing} \sft{},
even for all our ablation runs on continued pre-training.
We observe that, on some long-context tasks, \alexedit{performance gains only emerge} after \sft{}, which means that best design choices can differ before and after \sft{}.
\arrtgedit{We also check if the base model's short-context performance is preserved.}

Guided by our evaluation protocol, we run comprehensive experiments with Llama-3-8B~(8K original context window; \citealp{dubey2024llama}) to study each component of long-context \ctraining{}, including data mixture, data and length scaling, \inst{}, and many other design choices such as cross-document attention masking and position extrapolation. 
Many of our findings are surprising or contradictory to existing claims, for example, (1) training only on long data hurts long-context performance, 
(2) training on longer sequences than the evaluation length helps, and (3) \sft{} on only short instruction data is sufficient for good long-context performance.
We outline our main takeaways and the structure of the paper in the takeaway box at the beginning of this section.

Our final model, \tf{ProLong}, achieves the best performance at a 128K context length among 10B-parameter models, while taking only $5\%$ of the data budget compared to Llama-3.1's long-context training \citep{dubey2024llama}. 
ProLong has a maximum context length of 512K tokens, making it one of the longest-context LMs available.\footnote{Throughout the paper, we use binary prefixes K$=2^{10}$, M=$2^{20}$, and B=$2^{30}$.}

\section{Guiding Model Development With Meaningful Evaluations}
\label{sec:eval}

A pre-requisite for training a strong LM is having a robust evaluation suite that can guide model development while tracking its utility in real-world applications.  
While synthetic benchmarks like needle-in-a-haystack \citep[NIAH;][]{gkamradt_llmtest_needleinahaystack_2024} and RULER \citep{hsieh2024ruler} have gained much popularity due to their simplicity and controllability, we are interested in a wider range of tasks that reflect practical usage, such as the ability to reason over the whole document.
In the following, we describe our evaluation protocols and 
showcase why they are critical to our model development.

\subsection{Evaluating on diverse and realistic tasks}
\label{subsec:eval_datasets}

We first make the decision to use \helmet{} \citep{yen2024helmet} as our evaluation suite,
as it is one of the most comprehensive long-context benchmarks.
\arrawedit{For fast iteration, we only use a subset of \helmet{} tasks for model development:}
\begin{itemize}[leftmargin=10pt]
    \setlength{\itemsep}{1pt}
    \setlength{\parskip}{0pt}
    \setlength{\parsep}{0pt}
  
    \item \textbf{Recall}: Given a JSON file with random key-values pairs, retrieve the value for a key.
    
    \item \textbf{RAG}: Answer a question given retrieved Wikipedia documents (\textit{NQ}, \textit{HotPotQA}, \textit{PopQA}).

    \item \textbf{Re-ranking}: Produce top-10 rankings from a  shuffled list of documents (\textit{MSMARCO}).

    \item \textbf{ICL}: Learn classification tasks from many in-context examples, where the \#classes ranges from 6 to 151; average of 5 datasets (\textit{TREC coarse/fine}, \textit{NLU}, \textit{Banking77}, \textit{Clinc-150}).

    \item \textbf{QA}: Answer a question given a full-length book (\textit{NarrativeQA).}

    \item \textbf{Summarization}: Summarize  long legal documents (\textit{Multi-LexSum}).
\end{itemize}

Overall, these diverse tasks reflect a range of long-context abilities including recall, reasoning, learning from context, and robustness to noisy inputs. 
\citet{yen2024helmet} also show that \helmet{} produces model performance trends \tgedit{that are more consistent with human perceptions} unlike other long-context benchmarks.
\arrawedit{
We evaluate the final model’s generalization using the remaining \helmet{} tasks, which were not involved in its development, 
and also report the final performance on other long-context benchmarks such as RULER \citep{hsieh2024ruler} and $\infty$Bench~\citep{zhang-etal-2024-bench} in \autoref{app:more_benchmarks}.
}

\begin{minipage}{0.53\textwidth}
  
  We showcase the importance of a robust evaluation suite in \autoref{tab:niah_vs_ours}.
  As a predecessor of our work, \citet{fu2024data} only consider needle-in-a-haystack (NIAH) and perplexity during model development; evaluations on 3 tasks from \helmet{} reveal major short-comings of their models despite perfect NIAH scores.
  We also see how NIAH and even the HELMET recall task become saturated for strong models (Llama-3.1-8B vs. 70B) while other task categories continue to detect differences in their long-context abilities.

  \end{minipage}
  \begin{minipage}{0.03\textwidth}
  $\;$
  \end{minipage}%
  \begin{minipage}{0.44\textwidth}
  \centering
  \small
  \vspace{-2pt}
  \captionof{table}{HELMET offers a more holistic long-context evaluation. %
  We reproduce \citet{fu2024data} on Llama-3-8B with \sft{}. We report the instruct Llama versions.
  } %
  \vspace{-8pt}
  \label{tab:niah_vs_ours}

  \setlength{\tabcolsep}{1pt}
  \renewcommand{\arraystretch}{1} 
  
  \begin{tabular}{lcccc}
  \toprule
  & & \multicolumn{3}{c}{\helmet{}} \\
  \cmidrule{3-5}
  Models & NIAH  & Recall & RAG & Re-rank \\ %
  \midrule

  \citet{fu2024data} & 100& \textcolor{red!60!black}{95.8} & \textcolor{red!60!black}{52.1} & \textcolor{red!60!black}{23.1}\\ %
  \midrule
  Llama-3.1-8B & {100} & 99.4 & 56.3 & 37.0\\ %
  Llama-3.1-70B & 100 & 100 & 62.1& 49.2\\ %

  \bottomrule
  \end{tabular}
\end{minipage}

We offer more details about the \helmet{} evaluation, including its careful choice of metrics, in \autoref{app:eval}.
If not otherwise specified, we average the performance for each category over all datasets and over evaluation lengths of 32K and 64K; 
for the final long-context score, we macro-average all categories.

\begin{minipage}{.62\linewidth}
\paragraph{Why not perplexity?} Besides synthetic recall tasks, many previous works rely on perplexity (PPL) for evaluating long-context extensions of LMs \citep{chen2023extending,fu2024data,lu2024controlled}, which is commonly measured on the PG19 books dataset~\citep{Rae2020Compressive}. 
We use the ablation experiment from \autoref{subsec:long_ratio} to showcase why perplexity  is not an indicative metric for developing long-context models. 
The experiment studies how the ratio of long documents  affects the performance. We report both our  evaluation and the perplexity measured on the last 32K tokens of 64K-length documents from PG19.
As shown in \autoref{fig:ppl_is_bad}, while using more long data continues to improve PPL, 
it is clear that using 100\% long data significantly hurts downstream long-context performance. 

\end{minipage}%
\begin{minipage}{.03\linewidth}
$\;$
\end{minipage}%
\begin{minipage}{.35\linewidth}
   \centering
  \includegraphics[width=0.99\textwidth]{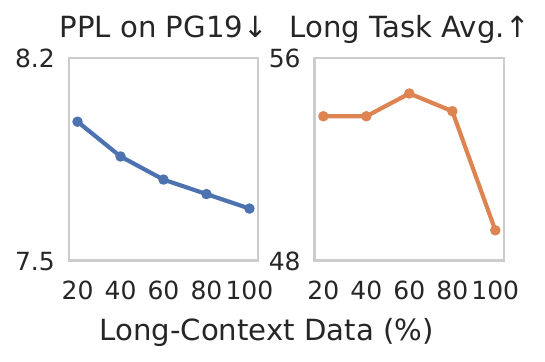}

   \captionsetup{width=.9\linewidth, hypcap=false} %
   \vspace{-1.8pt}
   \captionof{figure}{Making design decisions based on perplexity (PPL) is not optimal for long-context downstream tasks. 
   }
   \label{fig:ppl_is_bad}
\end{minipage}

\subsection{Evaluating after \inst{}}
\label{subsec:eval_after_sft}

\Inst{} (\sft{}; \citealp{ouyang2022training}) is an additional training stage that fine-tunes the model on a small amount of natural-language instructions and corresponding responses; it enables a base LM to address user queries in a chat format and has become a standard step for producing frontier LMs.
Here, we consider the difference between evaluating a model \textit{before} or \textit{after} \sft{}.

In preliminary experiments, we continue training \llamabase{} on 5B-token subsets from the data mix by \citet{fu2024data}. The mix is based on SlimPajama \citep{cerebras2023slimpajama} and upsamples long documents to constitute roughly 70\% of tokens, while retaining the original domain proportions.
Then we conduct \sft{} on several intermediate checkpoints with  UltraChat~\citep{ding-etal-2023-enhancing}.

\begin{wrapfigure}{r}{0.41\textwidth}
  \centering
  \vspace{-15pt}
  \includegraphics[width=0.4\textwidth]{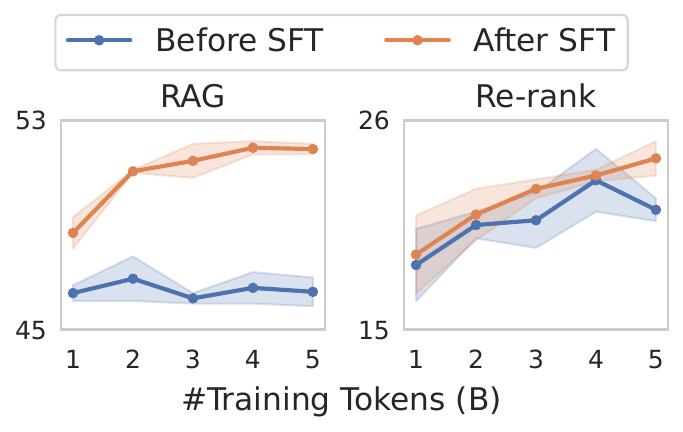}
  \caption{
      Some long-context improvements are only observed when evaluating after \sft{}. %
      We report the mean and std over two training runs.
  }
  \vspace{-10pt}
  \label{fig:small_fuyao_before_and_after_sft}
\end{wrapfigure}

We show the benchmarking results before and after \sft{} in \autoref{fig:small_fuyao_before_and_after_sft}. 
Long-context evaluation shows clearer signals when it is conducted after \sft{}:
(1) \sft{} shows that the model continues to improve with more training tokens on \rag{} and re-ranking, while the improvement is less clear or does not exist when evaluated before \sft{}.
(2) \sft{} enables evaluation on realistic applications like QA and summarization, which require instruction following and \arrtgedit{would otherwise fail completely}.
We also note that the variance from two random training runs is not substantially higher after the additional \sft{} phase. Therefore, unless otherwise specified, we report the long-context performance \textit{after} \sft{}.

We dive deeper into \inst{} in \autoref{sec:sft} and explore different training datasets, as well as the use of synthetic long instruction  data. However, we find that simply fine-tuning on UltraChat remains a surprisingly competitive choice.

\subsection{Checking that short-context performance is preserved}
\label{subsec:eval_short}

Long-context abilities should not come at the expense of short-context performance, particularly since short-context evaluations cover a wider range of capabilities, e.g., world knowledge, commonsense, and mathematical reasoning.
However, short-context evaluation has largely been neglected by previous long-context research.
We %
report on 5 tasks from the the Open LLM Leaderboard \citep{open-llm-leaderboard}: HellaSwag \citep{zellers2019hellaswag}, MMLU \citep{hendrycks2021measuring}, ARC-challenge \citep{clark2018think}, WinoGrande \citep{sakaguchi2021winogrande}, and GSM8K \citep{cobbe2021training}. 
We evaluate short-context performance \textit{before} \sft{}, \alexedit{since this allows for a direct comparison to  the base model which was used as initialization for the long-context training.}

\begin{minipage}{0.37\textwidth}
\paragraph{Previous techniques deteriorate short-context performance.}
We show in \autoref{tab:fuyao_short} that both training-free position extrapolation, as well as fine-tuning with an existing long data mixture \citep{fu2024data}
do not preserve the strong performance of \llama{} on standard short-context tasks.
This motivates us to find data sources which retain the initial model's strong short-context performance.

\end{minipage}%
\begin{minipage}{0.03\textwidth}
$\;$
\end{minipage}%
\begin{minipage}{0.58\textwidth}

\small
\centering

\setlength{\tabcolsep}{3pt}
\small

\captionof{table}{
Applying position extrapolation (PE)  to
 Llama-3-8B %
by changing the RoPE frequency base (\autoref{app:pos_extra})
or fine-tuning it
on a long-context SlimPajama mixture \citep{fu2024data, cerebras2023slimpajama} deteriorates the performance of this top-shelf pre-trained LM on short-context tasks.
}
\label{tab:fuyao_short}

\begin{tabular}{lcccccc}
\toprule
  & HSwag & MMLU & ARC-c & WG & GSM8K  \\
\midrule

\textit{Llama-3-8B} & 82.1 & 66.5 & 59.4 & 77.1 & 44.7  \\
+ PE & \textcolor{red!60!black}{81.5} & \textcolor{red!60!black}{64.7} & \textcolor{red!60!black}{58.1} & \textcolor{red!60!black}{75.5} & \textcolor{red!60!black}{40.1} \\
+ SlimPajama & \textcolor{red!60!black}{81.0} & \textcolor{red!60!black}{63.1} & \textcolor{red!60!black}{57.8} & \textcolor{red!60!black}{75.1} & \textcolor{red!60!black}{40.6} \\

\bottomrule
\end{tabular}%
\end{minipage}
\section{Long-Context Data Curation}

\label{sec:data}

The quality and composition of training data has been found to be the most important factor for LM pre-training \citep{penedo2023refinedweb, wettig2024qurating,li2024datacomplm} and is therefore
a primary focus of our study. %
To make data decisions, we perform ablation experiments: 
we continue to train \llamabase{} for 5B tokens with a maximum length of 64K tokens and evaluate according to \autoref{sec:eval}. See \autoref{app:how_to_ablate} for more details of our ablation setting.

We aim to boost the long-context task performance while preserving the short-context performance of the original model. %
Starting from the intuition that the data should be a mixture of long and short documents, we study these choices separately.
In our ablations, the long data is comprised of single-document chunks of 64K tokens, whereas for the short data, we construct batches by packing  documents until we reach 64K tokens per sequence.

\subsection{Code repositories and books are good sources of long-context data}
\label{subsec:whatlongdata}

\begin{minipage}{.69\linewidth}
     \paragraph{SlimPajama.} We analyze the quantity of long data in SlimPajama (SP; \citealp{cerebras2023slimpajama}). \autoref{tab:data_long_stats} shows that books account for the majority of long-context tokens. 
     When inspecting the long data in CommonCrawl (CC), we observe that though varied in quality, it also contains some book-like content, which future work could identify via data selection methods.

     \paragraph{Code repositories.} While only few files from GitHub reach a very long length (which also tend to be lower quality as suggested by \citealp{singh2024brevity}), we construct an abundant source of long-context data from the Stack \citep{kocetkov2023the} by concatenating all files from a repository to form a single document. Unlike \citet{guo2024deepseek}, we do not order the files based on dependencies, which should increase the distance between dependent files and reduce recency bias.
\end{minipage}%
\begin{minipage}{.03\linewidth}
$\;$
\end{minipage}%
\begin{minipage}{.28\linewidth}
    \centering
    \small
    \captionsetup{width=.9\linewidth, hypcap=false} %
    \captionof{table}{Long text documents ($\ge$64K tokens) by data sources.}
    \label{tab:data_long_stats}
    \begin{tabular}{lc}
    \toprule
    \multirow{2}{*}{Data} & \#Long \\
                          & tokens \\
    \midrule

    \textbf{Code Repos} & \textbf{98.8B} \\
    \textbf{SP/Books} & \textbf{33.2B} \\
    SP/CC & 15.3B \\
    SP/Arxiv & 5.2B \\
    SP/GitHub & 2.8B \\
    SP/Wiki & 0.1B \\
    SP/StackEx & $<$0.1B \\
    SP/C4 & $<$0.1B \\
    \bottomrule
    \end{tabular}

\end{minipage}

\paragraph{Data mixture.}
We train models with 60\% of long-context data %
and 40\% of our \shortmix{} (\autoref{sec:short_data}).
\autoref{tab:composition_ratios} shows that using code repositories alone performs the best on stress-test recall tasks. %
Meanwhile, books are more broadly beneficial for in-context learning, summarization and re-ranking. 
An equal mix of books and code repositories achieves the best overall performance.
Note that short-context task performance remains consistent due to our high-quality short data mix.

\begin{table}[h!]

    \small
    \centering
    \setlength{\tabcolsep}{4pt}
    
    \caption{Impact of different long data sources. Long-context performance is averaged over 32K and 64K lengths.}%

    \label{tab:composition_ratios}

    \vspace{-5pt}
    \begin{tabular}{lccccccccc}
    \toprule
    \multirow{2}{*}[-0.5ex]{Long Data (60\%)} & \multicolumn{7}{c}{Long-Context} & \multicolumn{1}{c}{Short-Context} \\
     \cmidrule(lr){2-8} \cmidrule(lr){9-9}
    & \jsonkv & \rag & \rerank & \icl & \qa & \summ & \avg & \avg \\

    \midrule
    CommonCrawl & 84.1 & 53.3 & 28.1 & 67.5 & \tf{35.2} & 37.0 & 50.9 & 66.5 \\
    \arrtgedit{ArXiv} & 90.3 & 51.8	& 28.0 & 68.0 & 33.7 & 36.7 & 51.4 & 67.5\\
    Books  & 94.9 & \tf{53.9} & \tf{30.7} & \tf{72.2} & 33.2 & \tf{37.7} & {53.8} & 65.5 \\%67.3 \\
    Code Repos & \tf{99.2} & \tf{53.8} & 29.0 &61.2  & 34.7 & 36.2 & 52.3 & {65.9} \\ %
    \midrule
    \arrtgedit{Books/Repos/ArXiv 1:1:1} & 98.3 & 53.9 & 29.4 & 66.9 & 35.5 & 35.5 & 53.3 & 66.9\\
    \rowcolor{orange!10!white} Books/Repos 1:1 & 96.0 & {54.9} & 29.4 & {73.9} & {35.7} & {37.9} & \tf{54.6} & {65.5} \\% \tf{67.5} \\
    \bottomrule
    \end{tabular}

\vspace{-10pt}
    \end{table}

\subsection{Training only on long data hurts long-context performance}
\label{subsec:long_ratio}

The ratio between short/long data is another crucial factor for downstream performance.
Prior work either trains only on long data \citep{peng2024yarn} or adds some  short training data \citep{yen-etal-2024-long, fu2024data}.
However, we are the first to systematically study the impact of  short/long ratio.

\begin{figure}[h]
    \centering
    \includegraphics[width=0.99\textwidth]{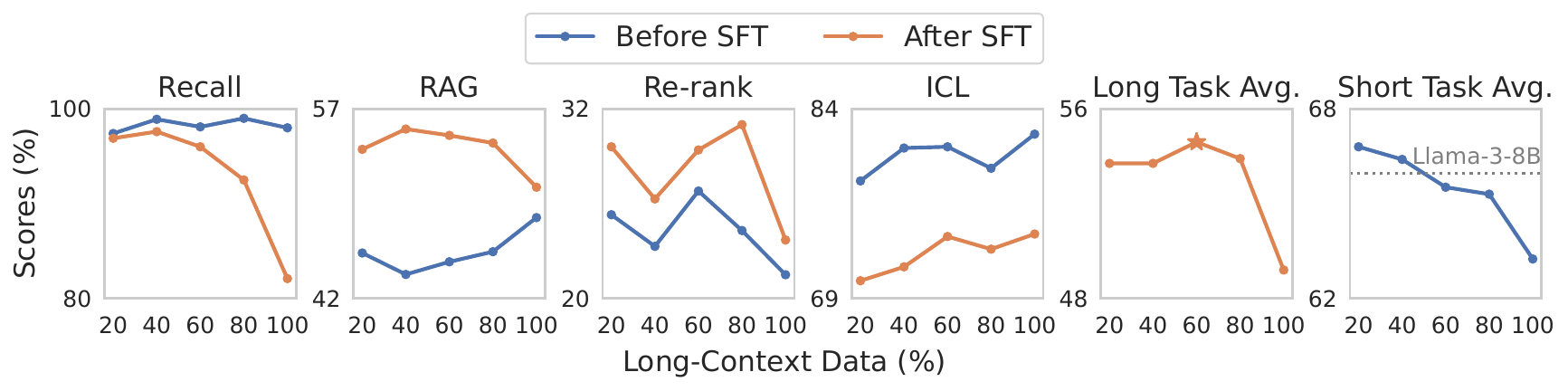}
    \vspace{-10pt}
    \caption{
    Impact of short/long data ratio.
    All models are trained on books/repos long data and our \shortmix{} for 5B tokens. 
    More long data initially improves long-context performance, but %
    then 
    becomes impairing. More long data also consistently degrades the short-context performance.
    }
    \label{fig:longratio}
    \vspace{-5pt}
\end{figure}

\autoref{fig:longratio} shows that
short task performance monotonically decreases as the long data increases.
The trends for long-context vary by tasks and are further complicated by \sft{}:
On tasks like recall and \rag{}, %
the performance before \sft{} prefers high proportions of long data, while the performance after \sft{} drastically deteriorates with more long data.
We hypothesize that specializing the model only on long data makes it a poor initialization for generic \sft{}---highlighting the importance of evaluating checkpoints after \sft{} (\autoref{subsec:eval_after_sft}).
While some long-context tasks benefit from more long data consistently (\icl{}) or show no clear pattern (re-ranking), the best average performance is achieved at 60\% long data and 40\% short data, which we adopt for our final \ours{} model.

\subsection{Choosing a high-quality short-context mix is important} \label{sec:short_data}

\begin{minipage}{0.72\textwidth}
We saw in \autoref{subsec:eval_short} that it is difficult to preserve the strong performance of Llama-3-8B on short-context tasks during long-context fine-tuning.
We adopt our best long-context settings (Book/repo data and 60\% long/40\% short) and study the impact of different short-context training mixes. We experiment with SlimPajama \citep{cerebras2023slimpajama}, FineWeb-Edu \citep{penedo2024finewebdatasetsdecantingweb}, DCLM-Baseline \citep{li2024datacomplm}, and our own ProLong \shortmix{}.
Our \shortmix{} is inspired by the ``stage 2 training'' in MiniCPM~\citep{hu2024minicpm} and Dolma-1.7 \citep{soldaini-etal-2024-dolma}, which use more knowledge-intensive, downstream-related data at the end of pre-training.
\autoref{tab:shortmix} shows the composition of our \shortmix{}.\footnotemark{}

\end{minipage}\footnotetext{
Since we do not truncate documents in the short data component unnecessarily, it includes a small percentage of documents longer than 8K. See \autoref{tab:shortmix_lengths} in the appendix for the dataset length statistics.}%
\begin{minipage}{0.03\textwidth}
$\;$
\end{minipage}%
\begin{minipage}{0.25\textwidth}
\centering
\small
\vspace{-2pt}
\captionof{table}{Our \shortmix{}.} %
\label{tab:shortmix}
\vspace{-5pt}
\begin{tabular}{lc}
    \toprule
    Components & \% \\
    \midrule
     FineWeb & 25 \\
     FineWeb-Edu & 25 \\
     Wikipedia & 10 \\
     Tulu-v2 & 10 \\
     StackExchange & 10 \\
     ArXiv & 10 \\
     OpenWebMath & 10 \\
     \bottomrule
\end{tabular}

\end{minipage}

\begin{table}[h!]

\small
\centering
\setlength{\tabcolsep}{3pt}

\caption{Impact of 
different short data sources.
The long-context performance is the average of 6 categories at the lengths of 32K and 64K. 
}
\label{tab:composition_short}

\begin{tabular}{lccccccc}

\toprule
\multirow{2}{*}[-0.5ex]{Short Data (40\%)} & \multicolumn{1}{c}{Long-Context} & \multicolumn{6}{c}{Short-Context} \\
 \cmidrule(lr){2-2}
 \cmidrule(lr){3-8}
 & Avg. & HellaS. & MMLU & ARC-c & WG & GSM8K  & Avg. \\
\midrule

\textit{Original model (Llama-3-8B)} & - & 82.1 & 66.5 & 59.4 & 77.1 & 44.7 & 66.0 \\
\midrule
SlimPajama & 52.9 & 81.2 & 63.0 & 58.5 & 76.2 & 41.9 & 64.2 \\
FineWeb-Edu   & 53.0 & 81.0 & 62.6 & 57.7 & 74.4 & 39.4 & 63.0 \\ %
DCLM-Baseline & 52.0& 82.0 & 65.6 & 59.6 & 77.4 & 39.4 & 64.8 \\ %
\rowcolor{orange!10!white} ProLong ShortMix &\tf{54.6}& 81.6 & 65.3 & 58.0 & 76.2 & 46.6 & \tf{65.5} \\ %

\bottomrule
\end{tabular}%

\end{table}

\autoref{tab:composition_short} demonstrates that the short data component has a substantial impact on both short-context and long-context downstream performance.
Our curated \shortmix{} outperforms other short data sources on both short and long-context tasks and our data domains are particularly important for retaining Llama-3-8B's performance on mathematical reasoning.
Surprisingly, we find that fine-tuning only using FineWeb-Edu---a dataset that is curated to help with knowledge-intensive tasks like MMLU---performs poorly as a short-context component, and we combine it with more diverse data sources in our \shortmix{}. 
DCLM-Baseline performs well on all short-context tasks except for GSM8K.
This can likely be improved by combining with math-related datasets, but as we added the DCLM-baseline ablation at the conclusion of the project, we leave this exploration to future work.

\arrawedit{
\paragraph{Comparison to prior efforts.}
To confirm the effectiveness of our long-context data curation, we conduct a head-to-head comparison  with the previous work by \citet{fu2024data}  in \autoref{app:compared_to_fuyao}. Our data mix significantly surpasses \citet{fu2024data} in both long-context and short-context tasks, underscoring the efficacy of \ours{}.
}

\section{Scaling the Size and Length of the Training Data}
\label{sec:howlong}

Training for more steps is well-known to improve downstream tasks in regular pre-training, but little analysis has been done in the context of long-context \ctraining{}.
We incorporate the lessons from our ablation experiments and arrive at the ProLong recipe, which we describe in detail in \autoref{sec:recipe}. Notably, we scale up the training budget to longer sequences (up to 512K) and more tokens (20B tokens at a maximum sequence length of 64K and an additional 20B tokens at 512K). We reset the learning rate schedule and increase the RoPE frequency base when switching from 64K to 512K context lengths.
In this section, we analyze the impact of these decisions. %

\begin{figure}[t]
    \centering
    \includegraphics[width=0.99\textwidth]{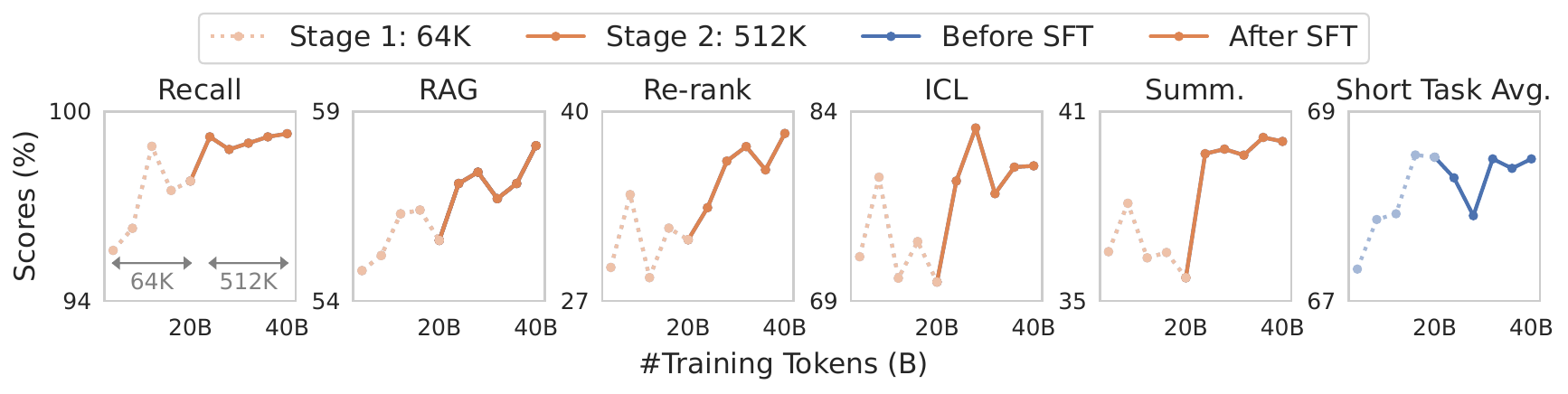}
    \caption{Performance (avg. of 32K and 64K) of our ProLong model throughout training. 
    }
    \label{fig:prolong_64k_512k}
    \vspace{-10pt}
\end{figure}

\paragraph{Increasing the number of steps helps.}
In \autoref{fig:prolong_64k_512k}, we plot the downstream performance of intermediate checkpoints of our 40B-token runs.
While the long-context performance fluctuates throughout training, we observe positive  trends on recall, \rag{}, re-ranking, and summarization.
For short-context tasks, we observe the average performance initially drops from the initialization, %
but gradually recovers. %
Performance again drops when switching from 64K to 512K sequence length, but also recovers with additional training.

\paragraph{Increasing the training length beyond the evaluation length helps.}
One might assume that we should train long-context models on the maximum sequence length that we want the model to support. Many works even emphasize extrapolation to even longer sequences at inference time \citep{press2022train, xiao2024efficient, xiao2024infllm, yen-etal-2024-long,chen2023extending}.
In contrast, we observe that training on a longer sequence length (512K tokens) substantially improves the long-context performance at a shorter evaluation length (64K tokens).

We establish this by %
initializing 
with a model that was trained for 20B tokens at 64K %
and either (1) continuing training at 64K, or (2) switching to the 512K training.
We use the same hyperparameters and data mixtures in either experiment.
We evaluate a checkpoint after 4B training tokens 
at a evaluation length of 64K.
Comparing the two runs
in \autoref{tab:64_vs_512},
we see consistent gains from switching to the 512K training length.\footnote{While we demonstrate the benefit of longer data, we note that training with longer sequences is more expensive, and may therefore not be the computationally optimal choice.}

\begin{table}[h!]

    \small
    \centering
    
    \caption{Impact of training models on different sequence lengths. 
    All the results are evaluated at a sequence length of  64K. 
    We see that training at a maximum length beyond the evaluation context window consistently improves the long-context performance.
    }
    \label{tab:64_vs_512}
    
    \begin{tabular}{lcccc}
    \toprule
    Max Seq. Length& \jsonkv & \rag & \rerank & \icl \\ %
    \midrule
    ProLong 64K training (20B) & 96.5 & 52.7 & 22.8 & 70.6 \\
    ~~+4B 64K training & 95.0 & 56.4 & 28.0 & 78.8 \\
    ~~+4B 512K training & \tf{98.5} & \tf{56.9} & \tf{32.9} & \tf{79.2} \\
    \bottomrule
    \end{tabular}

\end{table}

\arrawedit{
    What is the benefit of training on longer sequences than used during evaluation?
    Here is our hypothesis: 
    Assume that in order to solve a certain task (e.g., recall), a model has to be trained on examples of dependencies that span a distance of precisely $d$ tokens, 
    i.e., there is no generalization between dependencies of different lengths.
    Also assume that these dependencies are equally likely to occur at any position in a sequence.
    Then, a document of length $nd$ will have $(n-1)(d-1)$ more dependencies of distance $d$ than $n$ documents of length $d$.
    While these assumptions do not hold in practice, 
    this simplified model still provides intuition for our empirical findings.
}

\section{\INST{} for Long-Context LMs}
\label{sec:sft}

In this section, we study how to best enable long-context language models to follow instructions.
We focus on \inst{} on instruction datasets \citep{ouyang2022training} and
leave reinforcement learning and preference optimization for future work.

All our experiments in this section use the \ours{} base model, which was 
trained for 
40B tokens at a maximum sequence length of 512K. 
In comparison, open-source instruction data are very short, e.g., 
UltraChat \citep{ding-etal-2023-enhancing} conversations have 1.2K tokens on average and 4.1K tokens maximum.
To bridge this gap, several works~\citep{xiong2023effective,dubey2024llama,xiong2024artificial}
have proposed to generate long instruction data synthetically.

We consider three popular \sft{} datasets---UltraChat~\citep{ding-etal-2023-enhancing}, \mbox{Tulu-v2}~\citep{ivison2023camels}, ShareGPT\footnote{\url{https://huggingface.co/datasets/RyokoAI/ShareGPT52K}.}---and three sources of synthetic data: For \textit{synthetic QA}, we prompt  %
\llamainst{}
to generate a question-and-answer pair given a random chunk from a long document; %
we reuse the QA pairs for \textit{synthetic RAG} but we present 
a random list of chunks from the document to mimic retrieved passages; %
for \textit{synthetic summarization}, we generate summaries for long books via recursive summarization \citep{wu2021recursively}.
For all synthetic data, we write several templates, which we sample at random to increase diversity. More details can be found in \autoref{app:generate_synthetic_data}. 
We always use a combination of 40\% synthetic QA, 30\% synthetic RAG, and 30\% synthetic summarization in our synthetic instruction dataset.
The hyperparameters for the instruction tuning experiments can be found in \autoref{tab:recipe}.

\paragraph{Short-context instruction data yields strong long-context results.} %
We first establish that UltraChat outperforms Tulu-v2 and ShareGPT
in \autoref{tab:sft}.
We therefore use it when studying the ratio of synthetic long-context instruction data in \autoref{tab:sft_synthetic}.
Surprisingly, we find that adding synthetic data does not improve the performance on these very long-context tasks, and adding even as little as 1\% synthetic data hurts the performance in our setting. 
\arrtgedit{
We also verify that this phenomenon persists even when we use a  more powerful data generator, such as Llama-3-70B~(\autoref{app:sft_synthetic_70b}).}
Therefore, we use only short-context UltraChat data for \sft{} of our final \ours{} model.

\begin{table}[h!]

    \small
    \centering
    
    \caption{Effect of different ratios of synthetic SFT data (mixed with UltraChat). We report the 32K-and-64K-averaged performance except tasks marked with $^\dagger$, which are evaluated at 512K for stress testing. The number of percentage is based on \#tokens, not \#samples.}
    \label{tab:sft_synthetic}
    \vspace{-5pt}
    
    \begin{tabular}{ccccccccc}
    \toprule
    \% Synthetic Data & JsonKV$^\dagger$ & RAG & Re-rank & ICL & QA$^\dagger$ & Summ.$^\dagger$ & Avg. \\%& Avg. \\
    \midrule
    \rowcolor{orange!10!white}  0\% & 65.7 & 58.1 & 38.5 & 80.3 & 49.7 & 42.1 & \tf{55.7} \\%& 69.4 \\
    1\% & 61.5 & 57.0 & 38.3 & 80.8 & 45.3 & 41.5 & 54.1 \\%& 69.5 \\
    3\% & 62.0 & 56.4 & 37.9 & 80.6 & 44.8 & 39.5 & 53.5 \\%& 69.7 \\
    10\% & 70.3 & 55.5 & 36.1 & 80.6 & 41.7 & 39.4 & 53.9 \\%& 69.6\\
    50\% & 45.8 & 48.8 & 18.8 & 70.5 & 42.3 & 33.3 & 43.3 \\ 
    \bottomrule
    \end{tabular}

\end{table}

Why do our conclusions about synthetic data differ from previous work?
We offer the following hypotheses:
(1) Previous work like \citet{xiong2024artificial,bai2024longalign} %
may have  insufficient long-context training %
and the synthetic  data acts as additional long-context training data.
(2) Our instruction dataset  is much smaller compared to the private instruction data used for Llama-3.1 \citep{dubey2024llama}---it is possible that when using an extensive short instruction dataset, mixing in synthetic long data avoids the model from degenerating on long-context tasks.

\begin{table}[t!]
    \centering
    \small
    \caption{The training recipe for \ours{}.} %
    \vspace{-5pt}
\label{tab:recipe}
    \begin{tabular}{lll}
        \toprule
        \multicolumn{3}{c}{\tf{Continued Long-context Training}} \\
        \midrule

        \tf{Data} & \multicolumn{2}{l}{30\% code repos, 30\% books, 3\% textbooks, 37\% ShortMix} \\ \addlinespace
        & ShortMix: & \multicolumn{1}{l}{27\% FineWeb-Edu, 27\% FineWeb,} \\
        && 11\% Wikipedia, 11\% StackExchange, \\
        && 8\% Tulu-v2, 8\% OpenWebMath, 8\% ArXiv \\ \addlinespace

        \multirow{2}{*}{\shortstack[l]{\tf{Length} \\ \tf{Curriculum}}} & Stage 1 (64K):& Code repos, books, and textbooks at length 64K \\ \addlinespace %

         & Stage 2 (512K): & Code repos: 50\% at length 512K, 50\% at length 64K \\
        && Books: 17\% at length 512K, 83\% at length 64K \\ 
        && Textbooks at length 512K \\ \addlinespace %
        \tf{Steps} & \multicolumn{2}{l}{Stage 1: 20B tokens (2.2K H100 hours), $\;$ Stage 2: 20B tokens (12.2K H100 hours)} \\ \addlinespace
        \tf{Model} & Initialization: & \multicolumn{1}{l}{Llama-3-8B-Instruct (original RoPE base freq. $5\times 10^{5}$)} \\
        & RoPE: & \multicolumn{1}{l}{Stage 1: $8\times 10^{6}$, Stage 2: $\;$ $1.28\times 10^{8}$} \\
        & Attention: & Full attention with cross-document attention masking \\ \addlinespace
        \tf{Optim.} & \multicolumn{2}{l}{AdamW (weight decay = $0.1$, $\beta_1 = 0.9$, $\beta_2 = 0.95$)} \\
        & LR: & $1e-5$ with $10\%$ warmup and cosine decay to $1e-6$, each stage \\
        & Batch size: & 4M tokens for stage 1, 8M tokens for stage 2 \\

        \midrule
        \multicolumn{3}{c}{\tf{Supervised Fine-tuning (SFT)}} \\
        \midrule
        \tf{Data} & UltraChat \\ \addlinespace
        \tf{Steps} & 1B tokens \\ \addlinespace
        \tf{Optim.} & \multicolumn{2}{l}{AdamW (weight decay = $0.1$, $\beta_1 = 0.9$, $\beta_2 = 0.95$)} \\
        & \multicolumn{2}{l}{LR = $2e-5$ (cosine decay to $2e-6$), warmup = $5\%$} \\
        & \multicolumn{2}{l}{Batch size = 4M tokens} \\      
          \bottomrule
    \end{tabular}
    \vspace{-12pt}
\end{table}

\section{The ProLong Model: Recipe and Results}
\label{sec:recipe}

\subsection{Final recipe}
We summarize the training recipe for \ours{} in \autoref{tab:recipe}.
Our final model starts from the \llamainst{} model and is trained on 64K sequence length for 20B tokens. 
It is then further trained on 512K sequence length for 20B tokens (\ours{} base), which we achieve using sequence parallelism \citep{li-etal-2023-sequence}.
We obtain the final \ours{} model via \sft{} of the base model on UltraChat.
One small difference on the data mixture between our ablations and the final model is that we mix in 3\% high-quality textbooks~\citep{chevalier2024language}, as book-like data are shown to be beneficial for long-context (\autoref{subsec:whatlongdata}) and textbooks are  highly educational. This also slightly changes the proportions of ShortMix.
You can find more details about our data processing (\autoref{app:datapacking}) and the training stack (\autoref{app:stack}) in the appendix.

In the following, we elaborate on several carefully ablated design choices in our recipe.

\paragraph{RoPE frequency base tuning.} We find that changing the RoPE~\citep{su2021roformer} frequency base to achieve position extrapolation~\citep{xiong2023effective,dynamicntk} significantly improves long-context performance, even with a significant amount of training. \autoref{app:pos_extra} shows our ablation on the best RoPE base to use. While the original Llama models use a RoPE base of $10^5$, we use a base of $8\times 10^6$ for the 64K setting and $1.28\times 10^8$ for the 512K setting.

\paragraph{Disabling cross-document attention.} 
\citet{ding2024fewer} show that masking out attention across document boundaries improve model performance and this was also used during Llama-3 pre-training \citep{dubey2024llama}. 
In \autoref{app:docmask}, we show that disabling cross-document attention in \ctraining{} benefits both the short and long-context performance.
Disabling cross-document attention can also result in higher training throughput, which we describe in more detail in \autoref{app:stack}.%

\paragraph{Starting from \llamainst{}.} 
While we conduct all our long-context training ablations with the base model of \llama{},
we use Llama-3-8B-Instruct as the initialization for the final \ours{} model. 
\autoref{app:init} shows that while slightly improving the long-context performance, Llama-3-8B-Instruct significantly enhances the short-context performance.

\subsection{\ours{} performance}

\arrtgedit{
We first verify that \ours{} preserves the base model's short-context performance in \autoref{app:sft_short_context}.}
We then present the final \helmet{} evaluation results of \ours{} in \autoref{tab:main}.
\arrtgedit{We use all available \helmet{} tasks here and please refer to \citet{yen2024helmet} for more details.}
We compare to a number of frontier long-context LMs, namely MegaBeam\footnote{\url{https://huggingface.co/aws-prototyping/MegaBeam-Mistral-7B-512k}.}, Llama-3.1~\citep{dubey2024llama}, Qwen2~\citep{yang2024qwen2}, Phi-3~\citep{abdin2024phi}, Mistral-Nemo\footnote{\url{https://huggingface.co/mistralai/Mistral-Nemo-Instruct-2407}.}, Jamba-1.5~\citep{team2024jamba15}, Claude-3.5-Sonnet~\citep{anthropic2024claude}, Gemini-1.5~\citep{reid2024gemini}, and GPT-4o~\citep{achiam2023gpt}.

\ours{} outperforms all 10B-scale models on our long-context evaluation. Notably, \ours{} outperforms Llama-3.1-8B-Instruct 
on all categories except summarization. 
\ours{} achieves this with only 5\% of Llama-3.1's long-context  data budget (40B vs. 800B tokens).
We also showcase the strength of \ours{} with several QA examples in \autoref{tab:examples}. 
\arrtgedit{We also evaluate \ours{} on more long-context benchmarks, namely RULER~\citep{hsieh2024ruler}  and $\infty$Bench~\citep{zhang-etal-2024-bench} in \autoref{app:more_benchmarks}, which further verify the strength of \ours{}.}

\begin{table}[t!]
\centering
\small
\setlength{\tabcolsep}{4pt}

\caption{Our main evaluation results on \helmet{} (\citealp{yen2024helmet}) at 128K context length.
For all models, we use the corresponding instruction version.
\ours{} is  the best performing 10B-scale LMs.
\alexedit{The complete set of results can be found in} \S\ref{app:full_results}. \arrtgedit{Results on RULER and $\infty$Bench can be found in \autoref{app:more_benchmarks}.}
}
\label{tab:main}
\begin{tabular}{lccccccccc}
\toprule
Model & Max Len. & \jsonkv{} & \rag{} & \icl{} & \rerank{} & \qa{} & \summ{} & Cite & \avg{}  \\
\midrule
\rowcolor{orange!10!white} ProLong (8B)& 512K & \tf{98.8} & \tf{63.2} & \tf{86.5} & \tf{22.5} & \tf{43.9} & \tf{29.2} & {1.4} & \tf{49.4} \\
MegaBeam-Mistral (7B) & 512K & 89.6 & 57.0 & 86.2 & 14.7 & 37.3 & 28.9 & \tf{4.0} & 45.4 \\
Meta-Llama-3.1 (8B) & 128K & 95.2 & 59.5 & 83.9 & 14.0 & 43.2 & 27.0 & 2.9 & 46.5 \\
Qwen2 (7B) & 128K & 38.2 & 45.0 & 77.5 & 3.6 & 36.8 & 6.8 & 2.3 & 30.0 \\
Phi-3-small (7B) & 128K & 22.3 & 33.8 & 79.6 & 1.9 & 27.5 & 6.6 & 3.0 & 24.9 \\
Mistral-Nemo (12B) & 128K & 14.6 & 40.0 & 84.0 & 0.0 & 22.5 & 18.5 & 0.5 & 25.7 \\

\midrule

Jamba-1.5-Mini (12B/52B) & 256K & 90.0 & 57.3 & 91.0 & 14.6 & 54.2 & 18.1 & 3.1 & 46.9 \\
Meta-Llama-3.1 (70B) & 128K & 90.7 & 56.2 & 81.4 & 24.5 & 56.3 & 31.6 & 7.5 & 49.7 \\
Claude-3.5-Sonnet & 200K & 94.7 & 38.1 & 61.0 & 7.2 & 12.6 & 36.6 & 18.7 & 38.4 \\
Gemini-1.5-Pro & 2M & 91.0 & 71.1 & 79.4 & 59.7 & 59.6 & 46.4 & 43.6 & 64.4 \\
GPT-4o & 128K & 99.9 & 70.2 & 86.3 & 50.0 & 59.3 & 43.2 & 44.3 & 64.8 \\

\bottomrule
\end{tabular}

\end{table}

\begin{minipage}{0.67\textwidth}
    Since most existing models do not support more than 128K tokens, 
    to showcase \ours{}'s 512K context length,
    we stress test \ours{} on the QA and summarization tasks from 32K to 512K\footnotemark{}. %
    \autoref{tab:512k} shows that \ours{} continues to improve at a longer context window. %

    \end{minipage}
    \begin{minipage}{0.03\textwidth}
    $\;$
    \end{minipage}%
    \begin{minipage}{0.3\textwidth}
    \centering
    \small
    \vspace{-5pt}
    \captionof{table}{ProLong at 512K.} %
    \vspace{-8pt}
    \label{tab:512k}
 
    \setlength{\tabcolsep}{2pt}
    \renewcommand{\arraystretch}{0.8} 
    
    \begin{tabular}{rcccc}
    \toprule
    & 32K & 64K & 128K & 512K \\
    \midrule
    QA &31.7&	43.7& 46.7&	\tf{49.7} \\
    Summ & 40.4 & 	39.8& 	41.5&\tf{42.1} \\
    \bottomrule
    \end{tabular}
    
    \end{minipage}\footnotetext{
   In QA and summarization, we truncate the documents at the evaluation length; hence an effective long-context model should demonstrate better performance on longer lengths.}

\begin{table}[t!]
    \centering
    \small
    \setlength{\tabcolsep}{4pt}
    \caption{Results
    on the NoCha benchmark~\citep{karpinska2024one}.\tablefootnote{\url{https://github.com/marzenakrp/nocha}. \nocha{} has a private test set and all evaluation is done by the \nocha{} authors. Hence, we report models from \autoref{tab:main} that are also on the  \nocha{} leaderboard.}
    \tgedit{\ours{} is the} only model that achieves above-random performance in the $<$75K category and it consistently beats Llama-3.1. Different from the original \nocha{} leaderboard, we report the average accuracy over all test instances without filtering the test examples based on the model's context window lengths.}
    \label{tab:nocha}
    \begin{tabular}{lccccc}
    \toprule
    Model & Max Len. & $<$75K & 75K-127K & 127K-180K & $>$180K \\
    \midrule
    \rowcolor{orange!10!white} ProLong (8B)& 512K & \tf{28.4} & 17.0 & 13.1 & \tf{20.3} \\
    MegaBeam-Mistral (7B) & 512K & 19.8 & \tf{18.3} & \tf{17.5} & 15.6 \\
    Meta-Llama-3.1 (8B) & 128K & 17.3 & 16.4 & 0.0 & 0.0 \\
    Mistral-Nemo (12B) & 128K & 13.6 & 0.4 & 0.0 & 0.0 \\
    
    \midrule 
    Jamba-1.5-Mini (12B/52B) & 256K & 27.2 & 28.0 & 24.4 & 6.2 \\
    Meta-Llama-3.1 (70B) & 128K & 42.0 & 25.0 & 0.0 & 0.0 \\
    Gemini-1.5-Pro & 2M & 24.7 & 38.8 & 35.3 & 46.9 \\
    GPT-4o & 128K & 55.6 & 58.4 & 0.0 & 0.0 \\
    
    \bottomrule
    \end{tabular}
\vspace{-10pt}
\end{table}

Besides \helmet{}, we also evaluate our models on \nocha{}~\citep{karpinska2024one}---a claim verification dataset on 67 recently published English fictional books. 
We chose this dataset because (1) it minimizes the data contamination problem as all the books are unlikely to exist in the model pre-training data; (2) all the claims are written by human readers %
and require global reasoning. %
Each test instance contains two contradictory claims, and the models must correctly judge both to pass.

\autoref{tab:nocha} demonstrates the \nocha{} evaluation results.
Among 10B-scale models, \ours{} achieves the best accuracy on the extremely long test instances ($>$180K); on test instances %
$<$75K tokens, \ours{} significantly outperforms other models and is the only model that is better than random guessing (25\%). 
This further showcases the strength of our training recipe and the \ours{} model.

\section{Related Work}
\label{sec:related_work}

\paragraph{Adapting existing LMs for long contexts.}
Many works explore 
extending the LM context windows with minimal training, 
either by
position extrapolation~\citep{chen2023extending,peng2024yarn,chen2024longlora,ding2024longrope,liu20242,zhang2024extending,zhu2024pose,zhao2024longskywork,wu2024long,hu2024longrecipe}
or 
manipulating the attention patterns~\citep{chen2024longlora,xiao2024efficient,xiao2024infllm,bertsch2023unlimiformer,jin2024llm}.
\citet{yoshida2020adding,choromanski2021rethinking,chevalier-etal-2023-adapting}
instead explore the idea of compressing the long contexts into shorter forms.
However, \citet{fu2024data,lu2024controlled} show that using  full attention, applying simple position  extrapolation, and fine-tuning the model on long documents reach much stronger results. 

Llama 3.1~\citep{dubey2024llama} and Jamba~\citep{lieber2024jamba} achieve long-context capabilities by adding a long-context \ctraining{} stage between standard pre-training and \inst{}, which is the setting we follow.
\citet{fu2024data} study the data engineering for this setting and argue that 0.5B tokens of domain-balanced, length-upsampled data is sufficient for acquiring the long-context recall ability---which we show is not sufficient if a more holistic  evaluation is taken.
\citet{xiong2023effective,dubey2024llama,lieber2024jamba,xiong2024artificial,an2024make,bai2024longalign} also adopt synthetically-generated long data in the \sft{} stage; however,
we find that 
using standard, short-context instruction data achieves the best long-context results in our setting.

\paragraph{Efficient long-context architectures.}
There have been many efforts in designing more efficient architectures, 
for example, 
linear attention/RNNs~\citep{gu2023mamba,dao2024transformers,ma2022mega,sun2023retentive,peng-etal-2023-rwkv,yang2024gated},
and 
alternative attention architectures~\citep{rubin2023longrange,sun2024you,yen-etal-2024-long}.
However,
they often require training from scratch and many have the inherent limitations in terms of long-context recall \citep{jelassi2024repeat,arora2024simple}.
Recent works explore hybrid models \citep{waleffe2024empirical,lieber2024jamba} or distilling existing LMs into hybrid models \citep{wang2024mamba} and show promising results.

\paragraph{Long-context evaluation.}
Many benchmarks have been proposed for long-context evaluation
\citep{shaham-etal-2023-zeroscrolls,hsieh2024ruler,krishna-etal-2023-longeval,zhang-etal-2024-bench,an-etal-2024-l,bai-etal-2024-longbench}
There are works  studying particular aspects of long-context LMs as well, 
such as positional bias \citep{Liu2023LostIT},  
in-context learning~\citep{bertsch2024context,li2024long}, and book-length summarization~\citep{kim2024fables}.
In this work, we follow  \citet{yen2024helmet}
for its diverse application coverage and reliable evaluations.

\section{Conclusion}
\label{sec:discussion}

We study the problem of given a short-context pre-trained LM, how to 
most effectively continually pre-train and \sft{} the model to be long-context.
We conduct thorough ablations on each component and 
many of our findings contradict existing practices or beliefs. 
We use all the findings to produce \ours{}, a new state-of-the-art long-context LM. %
We release all our code, data, and models publicly 
and hope that our findings  will
boost research and applications of long-context LMs.

\subsubsection*{Limitations}
Although we aim to ablate the major components of our training recipe,
due to resource limitations, we cannot exhaust all aspects, such as the optimization hyperparameters and additional data mixtures.
We also limit ourselves to the 10B-scale regime and the Llama-3  models, which may limit the generalizability of our findings and recipe. 
Another concern is that we are overfitting to the tasks chosen for model development---however, we do not directly train on those datasets and guiding model development with benchmark tasks has become a common practice in pre-trained LM development. We also show that our final recipe and model perform well on  additional evaluation datasets, such as \nocha{}, RULER, $\infty$Bench, and held-out datasets from HELMET.

\subsubsection*{Acknowledgments}
We acknowledge Mengzhou Xia, Zexuan Zhong, Samyak Gupta, Dan Friedman, Yihe Dong, Abhishek Panigrah, Adithya Bhaskar, Colin Wang, Carlos Jimenez, and other members of Princeton Language and Intelligence for their helpful feedback and discussion. We also thank Luca Soldaini for providing comments on a draft.
We thank Marzena Karpinska, Tanya Goyal, and Mohit Iyyer for their help with the \nocha{} evaluation.
Tianyu Gao is supported by an IBM PhD Fellowship.
This work is gratefully supported by an NSF CAREER award (IIS-2239290), a grant from Intel, Cisco Research, and Microsoft Azure credits through the ``Accelerate Foundation Models Academic Research'' Initiative.

\bibliography{ref}
\bibliographystyle{colm2024_conference}

\appendix
\clearpage

\section{Experiment Details}

\subsection{Evaluation}
\label{app:eval}

\begin{table}[h!]
    \centering
    \small
    \caption{The details for our long-context evaluation following \helmet{} \citep{yen2024helmet}. }
\label{tab:eval_datasets_citation}
    \begin{tabular}{rlp{0.65\textwidth}}
        \toprule
        \tf{Category} & \tf{Metrics} & \tf{Tasks and Datasets} \\
        \midrule
        \tf{\jsonkv{}} & SubEM & Given a randomly-generated long JSON file and a key, retrieve the corresponding value~\citep{Liu2023LostIT}. \\ \addlinespace
        \tf{\rag{}} & SubEM & Given a question and many retrieved Wikipedia documents (shuffled), answer the question~\citep{Liu2023LostIT}. Datasets: \emph{NaturalQuestion}~\citep{kwiatkowski2019natural}, \emph{HotpotQA}~\citep{yang-etal-2018-hotpotqa}, and \emph{PopQA}~\citep{mallen-etal-2023-trust}.\\ \addlinespace
        \tf{\rerank{}} & nDCG@10 & Given a query and many retrieved documents (shuffled), re-rank the top-10 documents. Datasets: \emph{MSMARCO}~\citep{bajaj2016ms}. \\ \addlinespace
        \tf{\icl{}} & Accuracy & Datasets selected from \citet{bertsch2024context}: \emph{TREC coarse}, \emph{TREC fine}~\citep{hovy-etal-2001-toward}, \emph{NLU}~\citep{liu2021benchmarking}, \emph{Banking77}~\citep{casanueva-etal-2020-efficient}, and \emph{Clinc-150} \citep{larson-etal-2019-evaluation}.\\ \addlinespace
        \tf{\qa{}} & GPT-4o score & Given a book, answer the question. Datasets (\# tokens): \emph{NarrativeQA} (medium: 73K; max: 518K; \citealp{kocisky2018narrativeqa}). \\ \addlinespace
        \tf{\summ{}} & GPT-4o score& Summarize a given legal document. Datasets (\# tokens): \emph{Multi-LexSum} (medium: 90K; max: 5M; \citealp{multilexsum}) \\
        \bottomrule
    \end{tabular}

\end{table}

\autoref{tab:eval_datasets_citation} shows all the datasets we used for the long-context evaluation from \helmet{}~\citep{yen2024helmet}.
Note that we did not use all the datasets from \helmet{} for efficiency reasons and we also do not want to overfit to \helmet{}.
We highlight some of the evaluation protocol improvements that \helmet{} implemented compared to previous benchmarks here:

\begin{itemize}[leftmargin=15pt]
    \item \tf{Sufficient context lengths and fine-grained control}. \helmet{} can evaluate models at a context length of 128K tokens and beyond. 
    The evaluation protocol also allows for reporting results at different lengths, giving developers fine-trained controls for different needs of long contexts.
    \item \tf{Better synthetic recall tasks}. As shown in \helmet{}, needle-in-a-haystack~\citep{gkamradt_llmtest_needleinahaystack_2024} is mostly saturated because of its simplicity---the model only needs to find a needle in some irrelevant context. We instead use the more challenging JSON KV task, first proposed in \citet{Liu2023LostIT} and included in \helmet{}, where the model is required to find the corresponding value to a given key among a large JSON file. 
    \item \tf{Using class-balanced demonstrations and abstract labels for ICL}. To disentangle models' ability of learning from demonstrations from their pre-training bias of the task or the dataset label distribution~\citep{pan-etal-2023-context}, \helmet{} samples the same number of demonstrations for each class and uses number labels (\emph{1}, \emph{2}, ...) instead of natural-language labels (e.g., \emph{location}, \emph{description}, ...).
    \item \tf{Model-based evaluation for long-context QA and summarization}. Instead of using traditional metrics like ROUGE (which has shown to be poorly indicative of the real model performance: \citealp{deutsch-roth-2021-understanding,deutsch-etal-2022-examining,goyal2023newssummarizationevaluationera,chang2024booookscore}), \helmet{} uses model-based evaluations to compare the reference answer and the model output. For QA, \helmet{} uses GPT-4o to score the model output given the question and the reference answer at a 0-3 scale.
    For summarization, \helmet{} takes a similar approach as \citet{zhang-bansal-2021-finding,gao2023alce}: it first uses GPT-4o to decompose the reference summary  into atomic claims; then it uses GPT-4o to check whether each reference atomic claim is covered by the model output (recall) and whether each sentence in the model output is covered by the reference summary (precision). \citet{yen2024helmet} show that the model-based evaluation correlates with human perceptions significantly better than traditional metrics.
\end{itemize}

\subsection{Data processing}
\label{app:datapacking}

\paragraph{Data sources.}
We list all the data sources we have explored in our ablations and main experiments here: 
the Stack~\citep{kocetkov2023the},
SlimPajama~\citep{together2023redpajama,cerebras2023slimpajama},
FineWeb (we use the 2023-50 snapshot), FineWeb-Edu (we use a random sample)~\citep{penedo2024finewebdatasetsdecantingweb},
Tulu-v2~\citep{ivison2023camels},
OpenWebMath~\citep{paster2024openwebmath},
textbooks~\citep{chevalier2024language},
and Dolma~\citep{soldaini-etal-2024-dolma}.
The Books, StackExchange, and ArXiv data are from SlimPajama. The Wikipedia data are from Dolma.

\paragraph{Data filtering and packing.}
For the short training data and the SFT data, we randomly sample and concatenate the documents or conversations into 64K chunks. The last document for each chunk is truncated. The truncated part is used as the beginning for the next chunk for the short training data but is discarded for the SFT data.
For the long-context training data, we filter out the documents that are shorter than 64K; we do the same for the 512K setting, while making sure that the 64K documents packed to 512K length are distinct from the 512K documents.

\paragraph{Final data mixture.} 
For 512K length, we use a mix of 64K and 512K long data. For the ratio of 64K/512K data, we choose 50\%/50\% for code and 83\%/17\%, which are roughly chosen according to the natural availability of very long data, i.e., there are relatively fewer books of length 512K than code repositories. One benefit of retaining 64K-long documents is that we can process these without sequence parallelism and the associated communication overhead.
We use a slightly different long data mixture in our ablations~(\autoref{tab:shortmix}) and our main \ours{} experiment~(\autoref{tab:recipe}).
For the final model, we mix 3\% textbooks into the long-context training data. The textbooks are open-source resources from libretexts.org, collected and made available by \citet{chevalier2024language}.
We pre-process the data by concatenating chapters from the same text books, as well as books from the same subject areas. This results in extremely long sequences which we pack into contexts of either 64K or 512K tokens.
Though we do not have an ablation for adding this data due to limited resources, we believe that it should have a slight positive effect to the final model performance as textbooks are highly educational long-context data.

\begin{table}[h!]
    \centering
    \caption{\% Proportion of long documents for the short data components used in \autoref{tab:composition_short}. These statistics are computed after packing and truncation and therefore correspond to the document lengths as seen by the model. We highlight that the proportion of documents beyond 32K is below 1\% for \shortmix{}.}
    \label{tab:shortmix_lengths}
\begin{tabular}{lrrrr}
\toprule
 & \multicolumn{1}{c}{$>$4K} & \multicolumn{1}{c}{$>$8K} & \multicolumn{1}{c}{$>$16K} & \multicolumn{1}{c}{$>$32K} \\
\midrule
$\;$ FineWeb & 1.4 & 0.3 & 0.1 & 0.0 \\
$\;$ FineWeb-Edu & 2.8 & 0.8 & 0.2 & 0.0 \\
$\;$ Wikipedia & 1.6 & 0.4 & 0.0 & 0.0 \\
$\;$ Tulu-v2 & 0.0 & 0.0 & 0.0 & 0.0 \\
$\;$ StackExchange & 0.6 & 0.1 & 0.0 & 0.0 \\
$\;$ ArXiv & 85.7 & 64.0 & 30.3 & 7.6 \\
$\;$ OpenWebMath & 11.1 & 4.3 & 1.2 & 0.3 \\
\cmidrule{2-5}
\shortmix{} & 10.9 & 7.2 & 3.2 & 0.8 \\
\midrule
SlimPajama & 11.3 & 7.4 & 4.9 & 3.2 \\
FineWeb-Edu & 2.8 & 0.8 & 0.2 & 0.0 \\
DCLM-Baseline & 4.9 & 1.7 & 0.4 & 0.1 \\
\bottomrule
\end{tabular}

\end{table}

\subsection{Implementation details}
\label{app:stack}

\paragraph{Technical stack.} We use various open-source packages and tools for the \ours{} training and evaluation. 
We use PyTorch \citep{paszke2019pytorch} and Hugging Face transformers~\citep{wolf2020transformers} for the model training. 
We use mosaic-streaming \citep{mosaicml2022streaming} for loading and mixing the data
and FlashAttention 2 \citep{dao2024flashattention} for efficient  attention implementation. 
We implement sequence parallelism based on DeepSpeed-Ulysses \citep{jacobs2023deepspeed} across groups of 8 GPUs on the same node. We only perform distributed attention if it is necessary, i.e., only on sequences of 512K length.
For long-context evaluation, we use \helmet{}~\citep{yen2024helmet} and for short-context evaluation, we use lm-eval-harness \citep{eval-harness}.

\paragraph{Attention and batching.}
Since we do document masking in attention~(\autoref{sec:recipe}),
we use the variable-length attention implementation from FlashAttention 2~\citep{dao2024flashattention} to speed up long-context training: 
for sequences that are concatenations of multiple short documents,
instead of computing the full attention with masking, 
we instead compute the attention for each individual document.
Since the complexity of attention is quadratic to the sequence length, this improves the training speed.
However, the improvement is negligible in a distributed training setting with FSDP, since GPUs processing short sequence batches have to wait on other GPUs processing long sequences.
We therefore implement a smart batching algorithm:
In our setting, a gradient step usually consists of multiple gradient accumulation steps, where each device processes a smaller minibatch.
We sort all the minibatches per training step by the sum of the squared lengths of documents in the sequence.
This leads to more balanced sequence lengths across the GPUs and effective speedups, as can be seen in \autoref{tab:batching}, without affecting the gradient updates or loss during training. However, the efficiency gains are diminished when training with more GPUs, as this reduces the number of gradient accumulation steps.

\begin{table}[h]
    \centering
    \small
    \caption{Throughput per device of our ablation runs from \autoref{tab:docmask}, when training with 8 Nvidia H100 GPUs with FSDP. Our strategy of reordering minibatches is important for realizing the speed benefits from variable-length attention.}
    \begin{tabular}{ll}
        \toprule
         & \multicolumn{1}{c}{Throughput}  \\
         & \multicolumn{1}{c}{(tokens/s/GPU)} \\
         \midrule
         64K full attention &  $\quad$2770 \\
         \cmidrule(lr){1-1}
         Variable-length attention & $\quad$2780$_{(+0.4\%)}$ \\
         \rowcolor{orange!10!white} $\;$ + Minibatch reordering & $\quad$3095$_{(+11.7\%)}$ \\
         \bottomrule
    \end{tabular}
    \label{tab:batching}
\end{table}

\paragraph{Token-averaged loss.}
We found that in the \sft{} stage, 
the distribution of the training tokens (in \sft{}, the tokens from the instructions are masked out and the models are only trained on the responses) on each GPU device can be extremely imbalanced, especially when there is synthetic data (most tokens in a synthetic data instance are from the instruction). 
Conventional all-reduce loss in distributed training averages over the sequences instead of valid tokens, which skews the optimization and also our control over the domain proportions. 
Instead, we change the all-reduce loss to be the average over all valid training tokens.
\cite{bai2024longalign} implements their \sft{} loss in a similar way.

\subsection{The ablation setting}
\label{app:how_to_ablate}

For all our ablations, unless specified, we train the \textit{base model} of Llama-3-8B (instead of Instruct) on a 64K sequence length for 5B tokens, with the same hyperparameters as specified in \Cref{tab:recipe}.
We choose this context length, as it is the highest power of 2 value for which we can train without sequence parallelism.
By default, we use the same training data as the 64K \ours{} setting, except that we remove the textbooks and use the ShortMix proportions in \autoref{tab:shortmix}.
For \sft{}, we use the same settings as specified in \Cref{tab:recipe}.

\subsection{Generating synthetic \sft{} data}
\label{app:generate_synthetic_data}

We prompt \llamainst{} to generate the synthetic data and 
\autoref{tab:prompt_genqa} shows the prompt we used for generating the synthetic QA data for books.
We also write predefined templates and randomly sample one for each synthetic instance to increase the diversity, and \autoref{tab:template_qa_question} provides some examples.

\begin{table}[h!]
    \centering
    \small
    \caption{
        Prompts for generating synthetic QA data.
    }
    \label{tab:prompt_genqa}
    \begin{tabular}{>{\raggedright\arraybackslash\tt}p{0.95\textwidth}<{}}
        \toprule
            Given the following snippet of a book, ask a relevant question and provide the answer. The question and the answer should follow the following rules:\\\\
(1) The question should be specific enough that it can only be answered with the snippet. The question should also be interesting and intellectual enough that a curious reader of the book would ask about it.\\
(2) The question and the answer should be comprehensible given just the whole book without highlighting the snippet. With that being said, the question should NOT refer to the snippet directly (e.g., do NOT say things like "Question: given the conversation in the snippet, what ..."). The answer also should not mention "the snippet …" explicitly (assuming that the snippet is never provided), but it can copy the snippet content as a reference when answering the question.\\
(3) The answer should be concise but also should provide references to the book when needed. For example, “Wellington Yueh betrayed the Atreides, as the book mentioned, '...'". \\\\

*** Start of the snippet ***\\\\

\{sampled snippet\}\\\\

*** End of the snippet ***\\\\

Before generating the question and the answer, first reason about what this snippet is about. In your generation, stick to the following format:\\\\

Reasoning: this snippet is about ...\\
Question: ...\\
Answer: ... \\
        \bottomrule
    \end{tabular}

\end{table}
\begin{table}[h!]
    \centering
    \small
    \caption{
        Examples for question prompts and templates used for generating diverse synthetic QA data. We sample one question prompt and one template each time and combine them with the documents and the generated QA pairs to form a synthetic training example.
    }
    \label{tab:template_qa_question}
    \begin{tabular}{>{\raggedright\arraybackslash\tt}p{0.95\textwidth}<{}}
        \toprule
        \multicolumn{1}{c}{Example question prompts for synthetic QA data}\\
        \midrule
        Given the document, please answer the question.\\
        Here is a piece of text; answer the following question based on it.\\
        Please answer the question using the provided content.\\
        Based on the given passage, respond to the question.\\
        Read the snippet and answer the question that follows.\\
        Using the provided text, answer the following question.\\
        \midrule
        \multicolumn{1}{c}{Example templates for combining questions, answers, and contexts for synthetic QA data}\\
        \midrule
        \{prompt\}\textbackslash{}n\textbackslash{}n\{documents\}\textbackslash{}n\textbackslash{}nQuestion: \{question\}\\
        \{prompt\}\textbackslash{}n\textbackslash{}n==== document starts ====\textbackslash{}n\{documents\}\textbackslash{}n==== document ends ====\textbackslash{}n\textbackslash{}nQuestion: \{question\}\\
        \{prompt\}\textbackslash{}n\textbackslash{}n\{documents\}\textbackslash{}n\textbackslash{}n\{question\}\\
        \{prompt\} Question: \{question\}\textbackslash{}n\textbackslash{}n\{documents\}\\
        \{prompt\} \{question\}\textbackslash{}n\textbackslash{}n\{documents\}\\
        \{prompt\}\textbackslash{}n\textbackslash{}n\{question\}\textbackslash{}n\textbackslash{}n\{documents\}\\
        \bottomrule
    \end{tabular}

\end{table}

\section{More Ablations}

\subsection{Position extrapolation}
\label{app:pos_extra}

\citet{xiong2023effective,dynamicntk} show that 
changing the RoPE frequency base to a larger value in 
continual long-context pre-training or in inference time can improve the long-context performance. 
\citet{dynamicntk} suggests that one should scale the frequency base by a factor of $t^{\frac{d}{d-2}}$, where $t$ is the ratio between the target sequence length and the original LM length, and $d$ is the attention head dimension.

We conduct ablation studies, at both 64K (same as our standard ablation setting as specified in \autoref{app:how_to_ablate}) and 512K (starting from \ours{}-64K and training with the 512K data mixture for 5B tokens) sequence lengths, on what frequency bases we should use. 
\autoref{tab:64k_rope} and \autoref{tab:512k_rope} show the results.
We first see that using the original 500,000 frequency base from Llama-3 leads to significant performance degradation. While dynamic NTK suggests $4 \times 10^6$, we find that further scaling it to $8 \times 10^6$ leads to better performance. Similar, we see that when scaling the 64K model to 512K, while dynamic NTK suggests a $64 \times 10^6$ frequency base, much larger frequency bases ($128 \times 10^6$ and $256 \times 10^6$) lead to better performance.
We use $8 \times 10^6$ for 64K and $128 \times 10^6$ for 512K for our final \ours{} models.

\begin{table}[h!]

    \small
    \centering
    \setlength{\tabcolsep}{4pt}
    
    \caption{Ablation study on RoPE frequency base at a maximum training length of 64K. Dynamic NTK~\citep{dynamicntk} roughly suggests to use 4m as the frequency base.}
    \label{tab:64k_rope}
    \begin{tabular}{cccccccccc}
    \toprule
    \multirow{2}{1.5cm}[-0.5ex]{\centering RoPE Base \\ ($\times 10^6$)} & \multicolumn{7}{c}{Long-Context} & \multicolumn{1}{c}{Short-Context} \\
     \cmidrule(lr){2-8} \cmidrule(lr){9-9}
     & \jsonkv & \rag & \rerank & \icl & \qa & \summ & \avg & \avg \\
    \midrule
    0.5 & 25.8 & 37.0 & 4.4 & 73.8 & 17.5 & 16.3 & 29.1 & 65.0\\
    4.0 & 81.3 & 47.8 & 18.2 & 76.5 & 31.8 & 36.3 & 48.7 & 65.3\\
    \rowcolor{orange!10!white}  8.0 & 96.0 & 54.9 & 29.4 & 73.9 & 35.7 & 37.9 & \tf{54.6} & \tf{65.5}\\

    \bottomrule
    \end{tabular}

    \end{table}

\begin{table}[h!]

    \small
    \centering
    \setlength{\tabcolsep}{4pt}
    
    \caption{Ablation study on RoPE frequency base at a maximum training length of 512K. Dynamic NTK~\citep{dynamicntk} roughly suggests to use $64 \times 10^6$ as the frequency base.}
    \label{tab:512k_rope}
    \begin{tabular}{cccccccccc}
    \toprule
    \multirow{2}{1.5cm}[-0.5ex]{\centering RoPE Base \\ ($\times 10^6$)}  & \multicolumn{7}{c}{Long-Context} & \multicolumn{1}{c}{Short-Context} \\
     \cmidrule(lr){2-8} \cmidrule(lr){9-9}
     & \jsonkv & \rag & \rerank & \icl & \qa & \summ & \avg & \avg \\
    \midrule
    64 & 98.8 & 57.8 & 30.4 & 82.2 & 38.2 & 38.3 & 57.6 & 68.3 \\
    \rowcolor{orange!10!white}   128 & 98.8 & 57.4 & 30.7 & 80.0 & 40.4 & 38.8 & 57.7 & \tf{68.6} \\ 
     256 & 98.8 & 56.8 & 33.8 & 79.8 & 37.9 & 39.7 & \tf{57.8} & 68.4 \\

    \bottomrule
    \end{tabular}

    \end{table}

\subsection{Document masks}
\label{app:docmask}

We experiment whether to use document masks in attention in \autoref{tab:docmask}. Standard training concatenates multiple short documents into a single sequence (in our case, a 64K sequence), uses a special token to separate documents, and performs full attention over the whole sequence. When the document masks are used, we do not allow the attention to cross the document boundaries.
We find that using document masks in continual long-context training leads to both better long-context results and short-context performance.
For all our other ablations and  the main experiment, we use document masks.

\begin{table}[h!]

    \small
    \centering
    \setlength{\tabcolsep}{4pt}
    
    \caption{Impact of using document masks in attention.}
    \label{tab:docmask}
    \begin{tabular}{rccccccccc}
    \toprule
    \multirow{2}{*}[-0.5ex]{Attention} & \multicolumn{7}{c}{Long-Context} & \multicolumn{1}{c}{Short-Context} \\
     \cmidrule(lr){2-8} \cmidrule(lr){9-9}
     & \jsonkv & \rag & \rerank & \icl & \qa & \summ & \avg & \avg \\
    \midrule

    No doc masks & 97.4 & 53.6 & 20.4 & 76.6 & 37.2 & 36.3 & 53.6 & 64.9\\
    \rowcolor{orange!10!white}   Document masks & 96.0 & 54.9 & 29.4 & 73.9 & 35.7 & 37.9 & \textbf{54.6} & \textbf{65.5}\\

    \bottomrule
    \end{tabular}

    \end{table}

\subsection{Initialization}
\label{app:init}

We use the base model for \llama{} as the initialization for all our ablations to make sure the findings are generalizable and are not confounded by the Llama instruction tuning. 
However, for our final \ours{} model, we use \llamainst{} as the initialization to achieve the best performance.
We see in \autoref{tab:base_vs_instruct} (using the ablation setting from \autoref{app:how_to_ablate}) that using \llamainst{} as the initialization achieves slightly better long-context performance and much stronger short-context performance.

\begin{table}[h!]

    \small
    \centering
    \setlength{\tabcolsep}{3pt}
        \caption{
        Differences of using the base \llama{} model vs.  \llamainst{}.
    }
    \label{tab:base_vs_instruct}
    \begin{tabular}{rccccccc}
    
    \toprule
    \multirow{2}{*}[-0.5ex]{Base Model} & \multicolumn{1}{c}{Long-Context} & \multicolumn{6}{c}{Short-Context} \\
  \cmidrule(lr){2-2} \cmidrule(lr){3-8}
    & Avg. & HellaS. & MMLU & ARC-c & WG & GSM8K  & Avg. \\
    \midrule
    
    Llama-3-8B-Base & 54.6 & 81.6 & 65.3 & 58.0 & 76.2 & 46.6 & 65.5 \\
    \rowcolor{orange!10!white} Llama-3-8B-Instruct & \tf{55.0} & 80.8 & 66.1 & 58.5 & 75.6 & 57.7 & \tf{67.7} \\
    
    \bottomrule
    \end{tabular}%

    \end{table}

\subsection{Instruction-tuning datasets}
\label{app:sft}

Initialized from the \ours{} base model, we experiment with different public, short-context \sft{} datasets. 
All runs use the same \sft{} hyperparameters as specified in \autoref{tab:recipe}.
\autoref{tab:sft} shows that using UltraChat leads to the best overall results. 
Note that this does not necessarily mean that UltraChat is the best \sft{} dataset for all base models or applications.

\begin{table}[h!]

    \small
    \centering
    
    \caption{
        Ablations on using different short-context \sft{} datasets.
    We report the 32K-and-64K-averaged performance except tasks marked with $^\dagger$, which are evaluated at 512K for stress testing.
    }
    \label{tab:sft}
    \begin{tabular}{ccccccccc}
    \toprule
   \multirow{2}{*}[-0.5ex]{SFT Data} & \multicolumn{7}{c}{Long-Context} \\ %
    \cmidrule(lr){2-8} %
    & \jsonkv$^\dagger$ & \rag & \rerank & \icl & \qa$^\dagger$ & \summ$^\dagger$ & \avg \\%& \avg \\
    \midrule
    \rowcolor{orange!10!white}  UltraChat & 65.7 & 58.1 & 38.5 & 80.3 & 49.7 & 42.1 & \tf{55.7} \\ %
    Tulu v2 & 61.5 & 45.4 & 25.1 & 81.8 & 40.4 & 40.3 & 49.1 \\%& 67.5 \\
    ShareGPT & 40.5 & 47.5 & 26.7 & 79.6 & 42.7 & 34.4 & 45.2 \\%& 69.0 \\
    \bottomrule
    \end{tabular}

\end{table}

\subsection{Synthetic data with a stronger data generator}
\label{app:sft_synthetic_70b}

We observe that mixing in synthetic data generated by Llama-3-8B-Instruct does not help with the long-context performance.
To ensure that this is not due to the low quality of the synthetic data, 
we also experiment with a stronger data generator, Llama-3-70B-Instruct.
We demonstrate the results in \autoref{tab:sft_synthetic_70b} and verify that using a stronger data generator does not change the conclusion.

\begin{table}[h!]

    \small
    \centering

    \caption{Effect of different ratios of synthetic SFT data (mixed with UltraChat). We report the 32K-and-64K-averaged performance except tasks marked with $^\dagger$, which are evaluated at 512K for stress testing. The number of percentage is based on \#tokens, not \#samples. ``(8B)'' and ``(70B)'' indicate that the synthetic data are generated by Llama-3-8B-Instruct or Llama-3-70B-Instruct. 
    Even though using synthetic data from a stronger model leads to slightly better performance than using a weaker model, only using short-context SFT data still achieves the best result in our setting. 
    }
    \label{tab:sft_synthetic_70b}
    
    \begin{tabular}{lcccccccc}
    \toprule
    \% Synthetic Data & JsonKV$^\dagger$ & RAG & Re-rank & ICL & QA$^\dagger$ & Summ.$^\dagger$ & Avg. \\%& Avg. \\
    \midrule
    \rowcolor{orange!10!white}  0\% & 65.7 & 58.1 & 38.5 & 80.3 & 49.7 & 42.1 & \tf{55.7} \\%& 69.4 \\
    1\% (from 8B) & 61.5 & 57.0 & 38.3 & 80.8 & 45.3 & 41.5 & 54.1 \\%& 69.5 \\
    1\% (from 70B) & 64.7 & 57.3 & 37.4 & 78.4 & 47.0 & 40.8 & 54.2 \\
    3\% (from 8B) & 62.0 & 56.4 & 37.9 & 80.6 & 44.8 & 39.5 & 53.5 \\%& 69.7 \\
    3\% (from 70B) & 65.7 & 57.4 & 38.0 & 80.1 & 48.7 & 42.5 & 55.4 \\
    10\% (from 8B) & 70.3 & 55.5 & 36.1 & 80.6 & 41.7 & 39.4 & 53.9 \\%& 69.6\\
    10\% (from 70B) & 66.3 & 57.0 & 33.4 & 81.2 & 45.3 & 38.4 & 53.6 \\
    50\% (from 8B) & 45.8 & 48.8 & 18.8 & 70.5 & 42.3 & 33.3 & 43.3 \\ 
    50\% (from 70B) & 55.8 & 53.9 & 23.5 & 74.1 & 50.7 & 39.9 & 49.7 \\
    \bottomrule
    \end{tabular}

\end{table}

\subsection{Comparison to \citet{fu2024data}}
\label{app:compared_to_fuyao}

We show a head-to-head comparison to the data strategy of \citet{fu2024data} in \autoref{tab:compared_to_fuyao}.
We see that under a fair comparison, our data mix significantly outperforms \citet{fu2024data} on both short and long-context tasks.
The main difference of the two data strategies is that 
\citet{fu2024data} proportionally up-sample long documents in each domain with an arbitrary ratio; \ours{} uses a mix of short and long documents, where the ratio of the mix and the domains for the long documents are carefully ablated.

\begin{table}[h!]

    \small
    \centering
    \setlength{\tabcolsep}{4pt}
    
    \caption{
        Comparison between \citet{fu2024data} and our model. 
        For a fair comparison, 
        we use the same initialization (Llama-3-8B), 
        same amount of data (5B), 
        and same hyperparameters (\autoref{app:how_to_ablate}). 
        The \ours{} data mix significantly outperforms 
        \citet{fu2024data} on both short and long-context tasks.
    }

    \label{tab:compared_to_fuyao}

    \begin{tabular}{lcccccccccc}
    \toprule
    \multirow{2}{*}[-0.5ex]{Data} & \multicolumn{7}{c}{Long-Context (After SFT)} & \multicolumn{2}{c}{Short-Context (Avg.)} \\
     \cmidrule(lr){2-8} \cmidrule(lr){9-10}
    & \jsonkv & \rag & \rerank & \icl & \qa & \summ & \avg  & Before SFT & After SFT \\

    \midrule
    \citet{fu2024data} & 95.8 & 52.1 & 23.1 & 72.0 & 31.0 & 37.0 & 51.8 & 64.1 & 65.4\\
    \rowcolor{orange!10!white} Our data mix & \tf{96.0} & \tf{54.9} & \tf{29.4} & \tf{73.9} & \tf{35.7} & \tf{37.9} & \tf{54.6} & \tf{65.5} & \tf{67.5}\\
    \bottomrule
    \end{tabular}

    \end{table}

\subsection{Short-context performance after SFT}
\label{app:sft_short_context}

We demonstrate the detailed short-context performance of \ours{} after SFT in \autoref{tab:short_peformance_beforeafter_sft}.

\begin{table}[h!]

    \small
    \centering
    \setlength{\tabcolsep}{4pt}
    
    \caption{Short-context performance of our model \textit{after} SFT. We also report a baseline using Llama-3-8B as the initialization and data from \citet{fu2024data}, trained with 5B tokens. ProLong is initialized from Llama-3-8B-Instruct. 
    ``Llama-3-8B-Instruct + UltraChat'': for a more fair comparison to ProLong, we conduct SFT on top of Llama-3-8B-Instruct with UltraChat. 
    ProLong largely retraines the short-context performance of Llama-3-8B-Instruct except MMLU and GSM8K. 
    We hypothesize that the close-source instruction tuning data of Llama-3-8B-Instruct is heavily engineered to improve math and knowledge-intensive tasks, which we do not have access to.
    ProLong achieves comparable results to ``Llama-3-8B-Instruct + UltraChat'', which further demonstrates  that our data mix effective retains short-context performance.
    }
    \label{tab:short_peformance_beforeafter_sft}
    
    \begin{tabular}{lcccccc}
    \toprule
    Model & HellaSwag & MMLU & ARC-c & WinoGrande & GSM8K & Avg. \\
    \midrule
    Llama-3-8B + \citet{fu2024data} & 82.5 & 63.9 & 63.6 & 75.1 & 42.2 & 65.4 \\
    Llama-3-8B & 82.1 & 66.5 & 59.4 & 77.1 & 44.7 & 66.0 \\
    Llama-3-8B-Instruct + UltraChat & 82.1 & 65.1 & 64.3 & 75.5 & 60.7 & 69.5 \\
    \rowcolor{orange!10!white}  ProLong & 82.8 & 64.6 & 64.7 & 76.2 & 58.9 & 69.4 \\
    \midrule
    Llama-3-8B-Instruct & 78.5 & 67.0 & 60.8 & 74.2 & 68.5 & 69.8\\
    \bottomrule
    \end{tabular}

\end{table}

\subsection{Evaluation on more benchmarks}
\label{app:more_benchmarks}

We also evaluate \ours{} on more long-context benchmarks, namely RULER~\citep{hsieh2024ruler}  and $\infty$Bench~\citep{zhang-etal-2024-bench} in \autoref{tab:infbenchm_ruler}. 
As pointed out by \citet{yen2024helmet}, 
RULER and $\infty$Bench cannot reliably  reflect long-context performance as their domain coverage is narrow and their evaluation metrics are noisy---as a result, we see unintuitive trends such as Gemini-1.5-Pro and Llama-3.1 (70B) perform worse than Llama-3.1 (8B).
Regardless, our model still performs the best on $\infty$Bench among 10B-scale models.

\begin{table}[h!]

    \small
    \centering
    \setlength{\tabcolsep}{3pt}
    
    \caption{Results on RULER and $\infty$Bench at  128K. 
    As pointed out by \citet{yen2024helmet}, 
    RULER and $\infty$Bench cannot reliably  reflect long-context performance as their domain coverage is narrow and their evaluation metrics are noisy---as a result, we see unintuitive trends such as Gemini-1.5-Pro and Llama-3.1 (70B) perform worse than Llama-3.1 (8B).
    Regardless, our model still achieves the best performance on $\infty$Bench among all 10B-scale models. 
    }
    \label{tab:infbenchm_ruler}
    
    \begin{tabular}{lccccccccccc}
    \toprule
     \multirow{2}{*}[-0.5ex]{Model} & \multicolumn{1}{c}{RULER} &  \multicolumn{10}{c}{$\infty$Bench}  \\
     \cmidrule(lr){2-2}  \cmidrule(lr){3-12} 
 & Avg.  & MC & QA & Sum & Diag & Calc & Find & Number & PassKey & KV & Avg. \\
    \midrule

    \rowcolor{orange!10!white}  ProLong (8B) & {71.9} & 65.1 & 22.0 & 19.8 & 4.5 & 0.0 & 27.4 & 100.0 & 100.0 & 92.8 & \tf{48.0}\\

    MegaBeam-Mistral & 78.9 & 53.7 & 18.5 & 24.8 & 12.0 & 0.0 & 24.3 & 99.7 & 100.0 & 36.4 & 41.0 \\
    Meta-Llama-3.1 (8B) & 81.3 & 67.2 & 15.5 & 26.7 & 23.0 & 0.0 & 33.1 & 99.5 & 100.0 & 55.0 & 46.7 \\
    Qwen2 & 26.7 & 39.7 & 5.2 & 15.5 & 8.5 & 0.0 & 24.9 & 76.3 & 94.6 & 0.0 & 29.4  \\
    Phi-3-small & 72.6 & 71.6 & 8.4 & 24.0 & 20.0 & 0.0 & 31.7 & 100.0 & 100.0 & 19.6 & 41.7\\
    Mistral-Nemo & 22.7 & 31.9 & 16.8 & 14.3 & 5.5 & 0.0 & 1.4 & 36.6 & 62.7 & 0.0 & 18.8\\
    \midrule
     Jamba-1.5-Mini & 87.8 & 76.0 & 17.9 & 0.0 & 3.5 & 0.0 & 31.1 & 100.0 & 100.0 & 45.6 & 41.6\\
    Meta-Llama-3.1 (70B) & 75.8 & 75.5 & 23.3 & 31.3 & 18.0 & 0.0 & 43.1 & 99.7 & 100.0 & 2.6 & 43.7 \\
    GPT-4o-mini & 80.8 & 78.2 & 19.1 & 24.8 & 21.5 & 0.0 & 69.7 & 100.0 & 100.0 & 80.4 & 54.9\\
    GPT-4o & 93.3 & 86.5 & 26.0 & 21.5 & 51.0 & 0.0 & 58.9 & 100.0 & 100.0 & 99.8 & 60.4\\
    Gemini-1.5-Pro & 65.3 & 77.5 & 27.7 & 29.0 & 97.5 & 0.0 & 58.0 & 100.0 & 100.0 & 70.4 & 62.2\\

    \bottomrule
    \end{tabular}

\end{table}

\section{Full Evaluation Results}
\label{app:full_results}

We provide all the evaluation results, including specific numbers for each dataset and evaluation length, at \url{https://shorturl.at/pULbx}.

\section{Examples}
\label{app:examples}
\autoref{tab:examples} shows some long-document QA examples with \ours{}, Llama-3.1, and GPT-4o.

\definecolor{mygreen}{RGB}{56,118,29} %

\begin{table}[h!]
    \centering
    \small
    \caption{
        Long-document QA examples with \ours{} (512K), Llama-3.1-8B-Instruct (128K), and GPT-4o (via the ChatGPT interface). Note that 
        it is very likely that the ChatGPT-version GPT-4o uses a retrieval component for long documents, hence it can process contexts beyond its 128K context window.
        We see that \ours{} can answer questions effectively with a 300K document while both Llama-3.1 and GPT-4o fail. 
        On a shorter document (Llama-3.1 paper), \ours{} can understand very technical questions and answer them accurately.
    }
    \label{tab:examples}
    \begin{tabular}{ll}

        \toprule
        \multicolumn{2}{c}{\ttt{Instacart S1 file (300K tokens)}} \\
        \midrule
        \multirow{3}{*}[1.2ex]{\parbox{0.3\textwidth}{\tf{Q}: How many shares and at what price did instacard sell their series G  redeemable convertible preferred  stock in June 2020 and July 2020?}} & 
        \parbox{0.63\textwidth}{\tf{\ours{}}: Instacart sold an aggregate of {\color{mygreen} 6,757,893} shares of their ... at a purchase price of {\color{mygreen}\$48.0919} ...
        } \\ \addlinespace
        & \parbox{0.63\textwidth}{\tf{Llama-3.1-8B}: {\color{red}The following is a II/5 [collapsed]}} \\ \addlinespace
        & \parbox{0.63\textwidth}{\tf{GPT-4o}: ... The company sold a total of {\color{red}6,758,000} shares at a price of {\color{mygreen}\$48.0919} per share during this period. } \\ 
        \midrule
        \multicolumn{2}{c}{\ttt{Llama-3.1 paper (94K tokens)}} \\
        \midrule
        \multirow{3}{*}[0ex]{\parbox{0.3\textwidth}{\tf{Q}: During training of the llama3 model, how many unexpected interruptions are caused by hardware related reasons?}} & 
        \parbox{0.63\textwidth}{\tf{\ours{}}: Approximately {\color{mygreen}78\%} of unexpected  ...} \\ \addlinespace
        & \parbox{0.63\textwidth}{\tf{Llama-3.1-8B}: According to Table 5, approximately {\color{red}58.7\%}  ...} \\ \addlinespace
        & \parbox{0.63\textwidth}{\tf{GPT-4o}: ... approximately {\color{mygreen}78\%} of unexpected interruptions  ...} \\ \addlinespace \addlinespace \addlinespace
        \multirow{3}{*}[2.5ex]{\parbox{0.3\textwidth}{\tf{Q}: Is Mark Zuckerberg a contributor to the paper?}} & 
        \parbox{0.63\textwidth}{\tf{\ours{}}: There is {\color{mygreen}no} mention of Mark Zuckerberg as a contributor to the paper.} \\ \addlinespace
        & \parbox{0.63\textwidth}{\tf{Llama-3.1-8B}: Yes, Mark Zuckerberg is {\color{red}listed as a contributor} ...} \\ \addlinespace
        & \parbox{0.63\textwidth}{\tf{GPT-4o}: Mark Zuckerberg is {\color{mygreen}acknowledged} in the paper for ..., but he is {\color{mygreen}not listed  as a core contributor} ...} \\ \addlinespace \addlinespace \addlinespace
        \multirow{3}{*}{\parbox{0.3\textwidth}{\tf{Q}: What ratio did the llama 3 team use to mix synthetic long-context data with short data in SFT?}} & 
        \parbox{0.63\textwidth}{\tf{\ours{}}: The llama 3 team used a ratio of {\color{mygreen}0.1\%} synthetic ...} \\ \addlinespace
        & \parbox{0.63\textwidth}{\tf{Llama-3.1-8B}: {\color{mygreen}0.10\%}} \\ \addlinespace
        & \parbox{0.63\textwidth}{\tf{GPT-4o}: The Llama 3 team used a mix ratio of {\color{red}30\%} synthetic ...} \\ \addlinespace \addlinespace \addlinespace
        \multirow{3}{*}[3.5ex]{\parbox{0.3\textwidth}{\tf{Q}: What benchmark did they evaluate on for "instruction-following" ability? Provide references too.}} & 
        \parbox{0.63\textwidth}{\tf{\ours{}}: They evaluated on {\color{mygreen}IFEval} (Zhou et al., 2023) for instruction-following ability. Reference: {\color{mygreen}Jeffrey Zhou, ... Instruction-following evaluation for large language models. arXiv preprint ...}} \\ \addlinespace
        & \parbox{0.63\textwidth}{\tf{Llama-3.1-8B}: The benchmark they evaluated on for "instruction-following" ability is {\color{mygreen}IFEval}. Reference: {\color{orange}Section 5.2.2, "General Knowledge and Instruction-Following Benchmarks"}} \\ \addlinespace
        & \parbox{0.63\textwidth}{\tf{GPT-4o}: The benchmark they used to evaluate "instruction- following" ability is {\color{mygreen}IFEval}. This is mentioned {\color{orange}in the context of assessing how Direct Preference Optimization (DPO) ...}} \\ \addlinespace
        \bottomrule
    \end{tabular}

\end{table}

\end{document}